\documentclass{article}
\usepackage{times}
\usepackage{cite}
\usepackage{amsmath,amssymb,amsfonts}
\usepackage{graphicx}
\usepackage{textcomp}
\usepackage{algorithmicx}
\usepackage{algpseudocode}
\usepackage{mathrsfs}
\usepackage{xcolor}
\usepackage{makecell}
\usepackage{multirow}
\usepackage{hyperref}
\usepackage{mathrsfs}
\usepackage{bbm}
\usepackage{color}
\usepackage{tabularx}
\usepackage{threeparttable}
\usepackage{circledsteps}
\usepackage{xspace}
\usepackage{float}
\usepackage{booktabs}
\usepackage{wrapfig}
\usepackage{subcaption}
\usepackage{mwe}
\usepackage{threeparttable}

\usepackage{bbding}
\usepackage{algorithm}
\usepackage{caption}
\usepackage[misc]{ifsym}
\usepackage{multicol}
\usepackage[preprint]{corl_2024} 

\author{
  Chengzhong Ma, Houxue Yang, Hanbo Zhang, Zeyang Liu, Chao Zhao, \\
  \textbf{Jian Tang, Xuguang Lan*, Nanning Zheng}\\
  \\
  Xi'an Jiaotong University\\
  $^{*}$ Correspondence to Xuguang Lan {\texttt{xglan@mail.xjtu.edu.cn}}\\
  Videos at - https://exdiff.github.io/index.html \\
}

\title{DexDiff: Towards Extrinsic Dexterity Manipulation of Ungraspable Objects in Unrestricted Environments}

\begin{document}
\maketitle

\begin{abstract}
Grasping large and flat objects (e.g. a book or a pan) is often regarded as an \textit{ungraspable} task, which poses significant challenges due to the unreachable grasping poses.
Previous works leverage \textit{Extrinsic Dexterity} like walls or table edges to grasp such objects. 
However, they are limited to task-specific policies and lack task planning to find pre-grasp conditions. This makes it difficult to adapt to various environments and extrinsic dexterity constraints.
Therefore, we present DexDiff, a robust robotic manipulation method for long-horizon planning with extrinsic dexterity. 
Specifically, we utilize a vision-language model (VLM) to perceive the environmental state and generate high-level task plans, followed by a goal-conditioned action diffusion (GCAD) model to predict the sequence of low-level actions.
This model learns the low-level policy from offline data with the cumulative reward guided by high-level planning as the goal condition, which allows for improved prediction of robot actions.
Experimental results demonstrate that our method not only effectively performs ungraspable tasks but also generalizes to previously unseen objects.
It outperforms baselines by a 47\% higher success rate in simulation and facilitates efficient deployment and manipulation in real-world scenarios. 
Videos at - https://exdiff.github.io/index.html
\end{abstract}


\section{Introduction}

Large and flat objects (e.g. books, plates, pans, etc.) are ubiquitous in our daily lives. However, robots fail to grasp such objects when they are lying on a surface due to size and pose constraints. To grasp such objects, proper grasps are always located on the side, which is always unreachable and hence makes objects \textit{Ungraspable}~\cite{ungraspable}. Humans typically position one side of the target object overhang by leveraging environmental constraints, or namely, extrinsic dexterity \cite{extrinsic} (e.g. push the target against the wall) before proceeding to grasp it. However, it remains an open question: How can robots utilize such extrinsic dexterity to robustly grasp large and flat objects in unrestricted environments like humans?

\begin{wrapfigure}{r}{0.5\textwidth}
    \centering
    \includegraphics[width=0.5\textwidth]{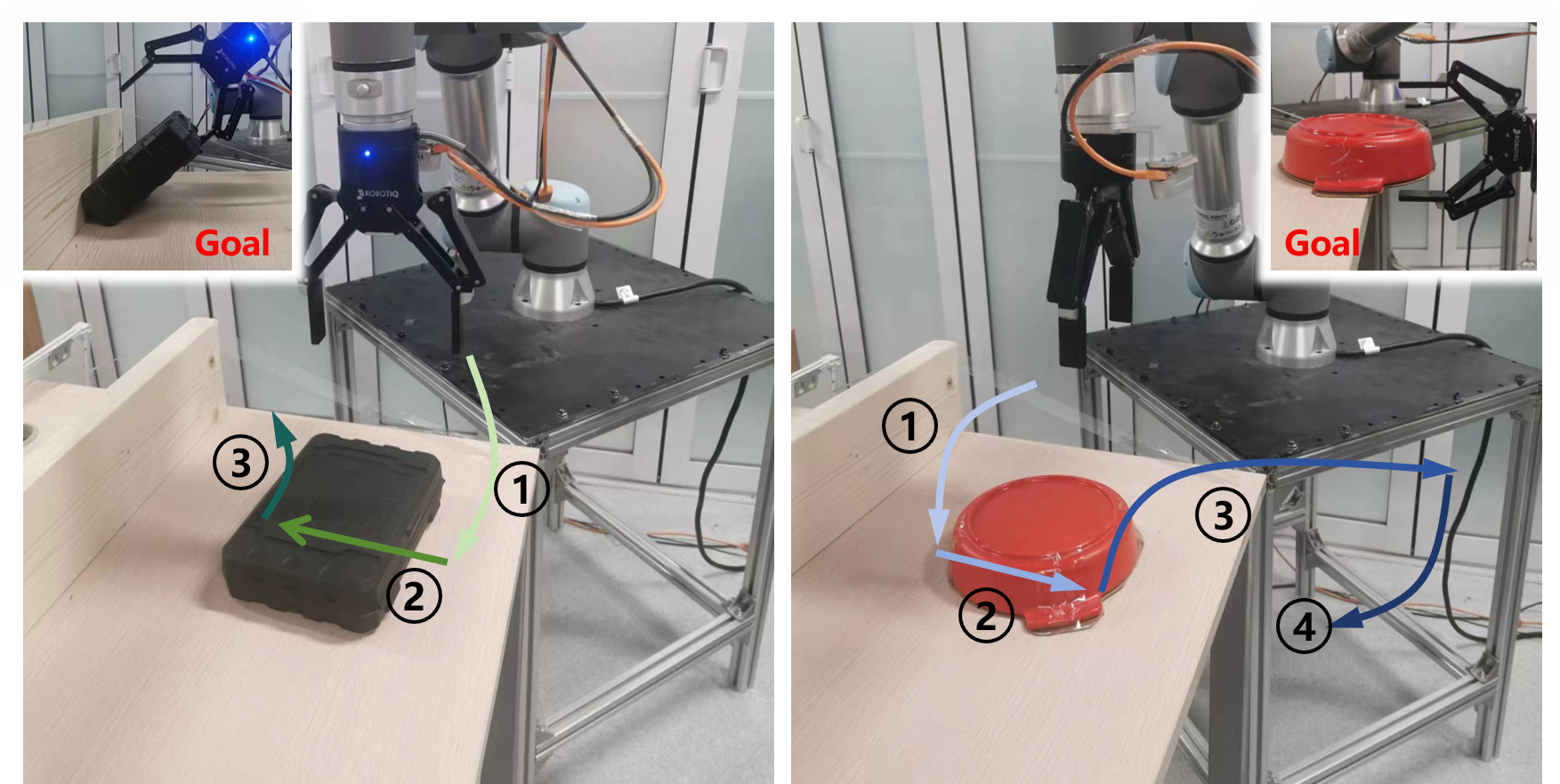}
    \caption{The robot may not grasp large flat objects on a tabletop from the top down. With the help of extrinsic dexterity, high-level task plans can be realized: \textbf{[Left]} Push the object against the wall, then rotate and grasp it from the side. \textbf{[Right]} Push the object to the edge of the table to keep it hanging and grasp it from the side.}
    \label{task}
    \vspace{-0.2cm}
\end{wrapfigure}

There are different types of extrinsic dexterity depending on the various external structures available in the observed environments (Figure.~\ref{task}), for example: (1) Push one side of the object against a wall, lift the other side by friction, and then grasp~\cite{ungraspable, sun2020learning, liang2021learning}, (2) Push the object to the edge of a table and then grasp it from the suspended side~\cite{pag, cheng2023enhancing, cheng2021contact}. To adapt to various environments and constraints, robots need to smartly plan for pre-grasping conditions based on any utilizable extrinsic dexterity and robustly execute low-level actions for grasping varied objects. 
Existing works are limited to specific extrinsic dexterity scenarios, relying on human-designed task and action planning. They lack the perception and adaptability to unrestricted environmental conditions, making it difficult to be applied in real-world scenarios.

In this paper, we propose a new grasping method based on extrinsic dexterity, \textbf{DexDiff}, comprising two key components: (1) recognizing external structures and obtaining high-level task planning through a vision-language model (VLM), and (2) processing multi-modal inputs to guide the prediction of action sequences using our proposed \textbf{Goal-conditioned Action Diffusion (GCAD)} model. In the GCAD model, we embed observation and \textit{return-to-go} (cumulative rewards from high-level task planning) sequences into the transformer architecture~\cite{transformer}. These help to learn horizon-aware policies according to different environmental conditions automatically. We evaluate our method in four typical extrinsic dexterity simulation scenarios, and the results indicate that DexDiff robustly accomplishes ungraspable tasks with a 70\% average success rate. It also shows generality and robustness against unseen objects with different positions, sizes, shapes, and surface friction. Furthermore, our method can be effectively deployed in real-world robotic standard scenarios and everyday life scenarios, serving as a benchmark case for integrating physical robot operations and extrinsic dexterity studies. 

In summary, our contributions include: 
    (1) We propose a robotic manipulation method called DexDiff, which can perceive the environment, formulate task plans, and achieve robotic motion planning, addressing ungraspable problems through extrinsic dexterity generally.
    (2) Our proposed Goal-conditioned Action Diffusion (GCAD) action prediction model can enhance the accuracy and generalization ability of long-horizon motion planning for robots by utilizing goal conditions provided by high-level task planning.
    (3) DexDiff achieves a higher simulation performance than baselines and successfully enables deployment on the physical robot in both standard external structures and real-life scenarios. This outcome further validates the effectiveness of our method in addressing practical ungraspable problems.

\section{Related Works}
\textbf{Extrinsic dexterity manipulation}.
Unlike the conventional approach of isolating the robot and the target object, extrinsic dexterity considers the interactions among the robot, the object, and the external environment~\cite{extrinsic}. 
Extrinsic dexterity is often used for non-prehensile manipulations and ungraspable manipulations. The former~\cite{zhou2023hacman,kim2023pre, cho2024corn} aims to manipulate objects without grasping them (moving the object to the goal pose), using a robot end-effector with a fixed contact point. The latter~\cite{ungraspable,pag} is applied when the objects to be grasped are large-sized, have special shapes, or are specifically constrained, thereby requiring the use of extrinsic dexterity to achieve robust grasping.

Current research has demonstrated its applicability to non-prehensile manipulations. In HACMan~\cite{zhou2023hacman}, they propose an approach that generalizes to diverse object shapes
using end-to-end training. However, they are limited to 3D push primitives. Another research~\cite{kim2023pre} outputs diverse motions and effectively performs time-varying hybrid force and position control~\cite{bogdanovic2020learning}, by using the end-effector target pose and controller gains as action space. However, it has a limited generalization capability across shapes since they represent object geometry via its bounding box. CORN~\cite{cho2024corn} proposes a novel contact-based object representation and pretraining pipeline to tackle this to get around these problems and adopts a controller with end-effector subgoals with variable gains~\cite{bogdanovic2020learning, kim2023pre} as the action space, which allows to perform dense, closed-loop control of the object.

In ungraspable manipulation scenarios, there exists some prior work discussing extrinsic dexterity: (1) push the object against the wall, then rotate and grasp it from the side~\cite{ungraspable, sun2020learning, liang2021learning}, (2) push the object to the edge of the table to keep it hanging and grasp it~\cite{pag, cheng2023enhancing, cheng2021contact}. However, these methods are strictly limited to a single external structure type and limited scenarios, making it difficult for the robot to adapt to different environments and constraints. 
In contrast, our work shows general ungraspable manipulation under different objects and environment settings, involving more complex object interactions while also generalizing to various unseen objects and external structure scenarios.



\textbf{Large language model-based task planning}.
Task planning in robotics aims to generate high-level abstract plans or strategies to achieve complex objectives or goals in the environment. Conventional methods require explicit primitives and constraints and lack scalability to open environments and the task generality required for real-world operations~\cite{aeronautiques1998pddl, kootbally2015towards, thomason2020jointly, vallati20152014}.
Recent research on large language models (LLMs) has shown strong reasoning abilities in task planning. 
For example, given language instructions, LLMs can be prompted to generate high-level step-by-step plans for long-horizon tasks over symbolic abstractions of the task~\cite{huang2022language, saycan, wang2023describe}. 
LLMs can also be integrated with pre-trained policies or existing APIs~\cite{chatgptforrobot, liang2023code}, and play the role of a manager for long-horizon tasks.
Furthermore, they can be finetuned upon real-world robot data to establish connections among languages, visual observations, and actions for joint training~\cite{rt1, rt2, li2023vision, padalkar2023open}.
However, LLMs are limited in their ability to represent concepts in the text and often struggle with grounding, such as reasoning over shapes, physics, and constraints of the real world~\cite{tellex2020robots, gao2023physically}. To this problem, LLMs can be integrated into larger vision-language models~\cite{palm, VLP} for high-level planning. When trained on sufficient data, these models can respect the physical constraints observed in the image inputs, thereby generating more feasible plans~\cite{gao2023physically}.
In our work, we fine-tuned a pre-trained VLM with captured scenario data and human prior knowledge that allows for effective perception and task planning in standard and everyday-life extrinsic dexterity tasks.

\textbf{Learning-based motion planning}.
Recently, reinforcement learning (RL) is often used to deal with non-prehensile manipulations~\cite{zhou2023hacman, cho2024corn, kim2023pre} and ungraspable manipulations~\cite{pag, ungraspable} because they are efficient in learning robotic control policies~\cite{Khandate-RSS-23, Zhang-RSS-23, Li-RSS-23, huang2022training}. However, the RL has some inherent limitations, such as relying on manually designed motions or primitives~\cite{cruciani2018dexterous, cruciani2019dualarm}, limited generalization capabilities~\cite{liang2021learning}, or difficult reward design~\cite{zeng2018learning, xu2021efficient, kim2023pre}. 
To address these issues, many studies consider learning policies from demonstrations~\cite{Haldar-RSS-23, Li2-RSS-23, Liu-RSS-23, Zeng-RSS-23, Schuppe-RSS-23, Kostrikov-RSS-23}. Imitation learning (IL) methods have shown promising results in real-world robot tasks~\cite{zhang2018deep, mandlekar2020learning, zeng2021transporter, mandlekar2020iris}. However, they require high sample quality and are limited to fitting sample distributions. The conditioned action generation methods utilize conditional designs such as prompts or rewards to guide or constrain policy learning~\cite{decisiontransformer, xu2022prompting}. They may learn actions that are better than the optimal trajectory, but they often fall into local optimality~\cite{xu2022prompting}.
Another approach is to use diffusion models for decision making~\cite{ajay2022conditional}, which have been widely utilized in recent research~\cite{ho2020denoising, Yoneda-RSS-23, janner2022planning, Liu2-RSS-23, Reuss-RSS-23}. It can infer possible execution trajectories in a robot control environment~\cite{huang2023diffusion, wang2022diffusion}. Diffusion Policy~\cite{diffusion} learns the gradient of the action-distribution score function~\cite{NEURIPS2019_3001ef25} and iteratively optimizes with respect to this implicit gradient field during inference. It has achieved good results in long-sequence action prediction. 
In our work, we coordinate the VLM task planner and goal-conditioned action diffusion motion planner, enabling our approach to autonomously make appropriate plans based on observations and learn robust operational strategies, rather than using fixed strategies.

\section{DexDiff}
The overview of DexDiff is shown in Figure.~\ref{framework}. It comprises two key components: (1) the environmental perception and task planning module based on the VLM, and (2) the action prediction and motion planning module based on Goal-conditioned Action Diffusion. 
For each scenario, we use the VLM for environmental perception to generate semantic plans. Then, we employ a semantic parser to extract relational tuples from these plans, guiding low-level decision-making. In the action prediction and motion planning module, the configuration encoder (C-encoder) and visual encoder embed inputs into the transformer. Combined with the return-to-go embedding and diffusion step embedding, the network can output the predicted action sequence after K-step denoising diffusion process. Finally, the robot's operational space controller (OSC) robustly completes the manipulation.

\subsection{Perception and Task Planning based on Finetuned VLM}
Due to the varying features of target objects and extrinsic dexterity structures in ungraspable tasks, efficient perception and task planning are crucial for downstream robotic action planning. In our scenarios, physical factors such as the size, shape, and friction of target objects, different types of extrinsic dexterity, the relative distance between the target and external structures, and the ease of specific robotic manipulations all need to be considered. Traditional heuristic methods can make judgments and plans based on predefined rules, but they often fail to make human-like decisions when facing unrestricted constraints. 

\begin{figure}
    \centering
    \includegraphics[width= \textwidth]{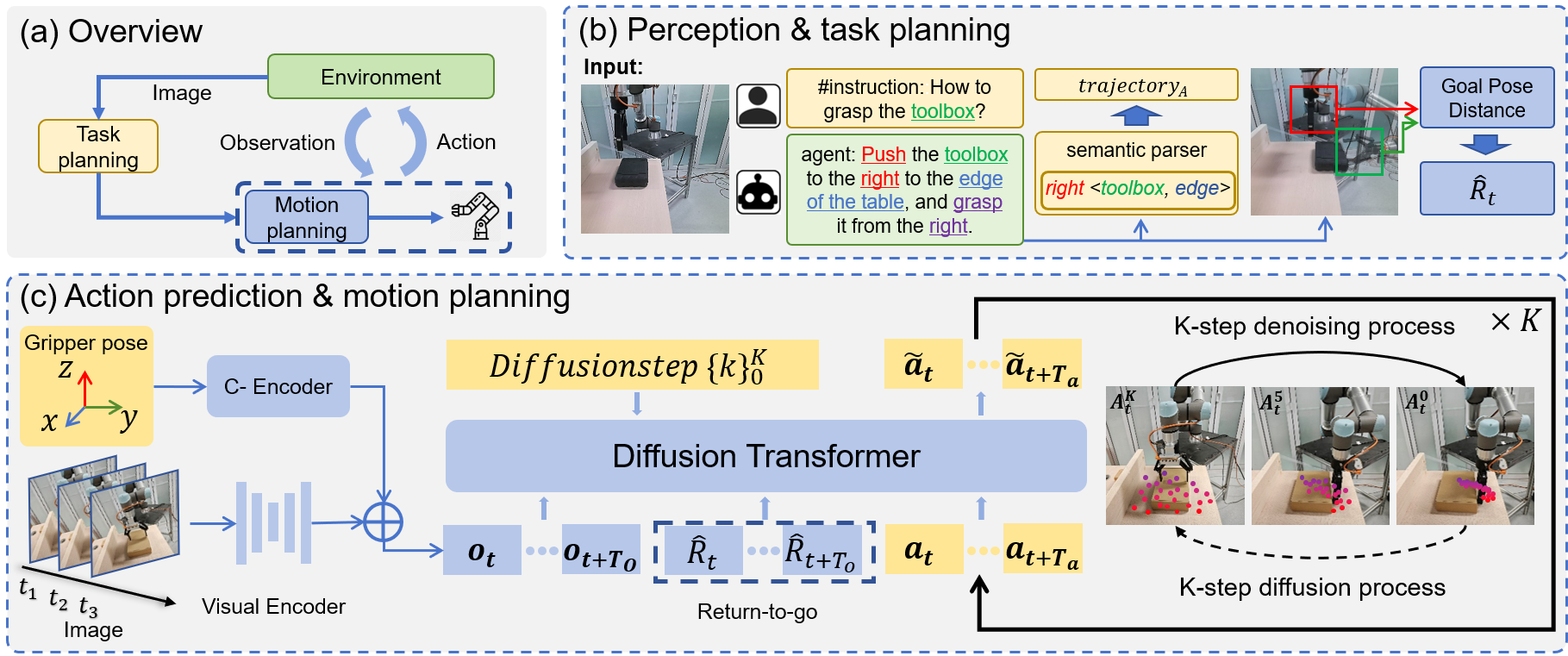}
    \caption{Our DexDiff method primarily consists of two modules: the high-level perception and task planning module based on the VLM and the low-level action prediction and motion planning module based on our GCAD model.
    }
    \label{framework}
    \vspace{-0.4cm}
\end{figure}

In DexDiff, a vision-language model called VisualGLM-6B~\cite{du2021glm} is finetuned to generate an initial environment perception and formulate a high-level task plan. We uniformly collected $\sim$ 1.6k images in the simulation scenarios and $\sim$ 1.4k images in the real scenarios, and provided a corpus of manipulation, including textual descriptions of standard robot skills, extrinsic dexterity structures, and manipulation directions, matched with corresponding scene images. 
To finetune the VLM, each sample $(E(obj), I, P(ski, obj, str, dir))$ consists of an initial expression $E(obj)$ to indicate the target object, a scene image of extrinsic dexterity $I$, and a labeled plan $P$ including texts of skills $ski$, external structures that can be utilized $str$, manipulation directions $dir$ and the target object $obj$. To be specific, we performed LoRA parameter fine-tuning for the pre-trained ViT-G visual encoder to deal with the image input $I$ and also performed LoRA parameter fine-tuning for the pre-trained ChatGLM~\cite{du2021glm} text encoder to deal with the instruction and planning. Then, we applied cross-attention by a query transformer and followed a simple cross-entropy loss of next-word prediction~\cite{du2021glm, xu2024sinvig}.

For example, as shown in Figure.~\ref{framework} (b), the perception and task planning module can confirm the target through the prompt inputs \# \textit{How to grasp the toolbox?} and reason about the appropriate task plan based on the external structures and spatial relationships observed in the image: \# \textit{Push the toolbox to the right to the edge of the table, and grasp it from the right.} 
From this sentence, we extract key information about extrinsic dexterity manipulation: ``Push to the right" (robot skill and relative direction of manipulation), ``edge of the table" (extrinsic dexterity structure), and ``grasp from the right" (goal grasp direction). These high-level task plans clarify the orientation of the external structure relative to the target object, guiding the direction of the initial action predictions and the final grasp pose for successful grasping.
Based on this, we extract relational tuples ($right < toolbox, edge >$) by a pre-trained semantic parser from task planning to selectively filter action sequences and trajectories. Then, we use goal-conditioned grasp poses to compute a series of return-to-go values, which are used to train the downstream GCAD model. By fine-tuning the pre-trained VLM model to learn the connections between the inputs, our approach can generalize the task planning to different scenes and target objects, as demonstrated in the experimental section of this paper.

\subsection{Goal-conditioned Action Diffusion}
We formulate the action prediction and motion planning with extrinsic dexterity as a Denoising Diffusion Probabilistic Model (DDPM)~\cite{ho2020denoising} for the first time and consider the use of upstream plan-guided return-to-go as the goal condition for predicting actions with observations (Figure.~\ref{framework} (c)). 

DDPM is a class of generation models in which output generation is modeled as a denoising process. In its setup, the diffusion model undergoes diffusion step $K$ iterations of denoising for the noise sample $x^K$ generated through Gaussian sampling. Then we get a series of actions with decreasing levels of noise, $x^k, x^{k-1},...,x^0$, the $x^0$ represents the desired output without noise, serving as the endpoint of the diffusion iterations. The diffusion iteration process follows the equation:
\begin{equation}
    x^{k-1} = \alpha(x^k-\gamma\varepsilon_\theta(x^k, k)+\mathcal{N}(0, \sigma^2I)),
    \label{diffusion-iterations}
\end{equation}
where $\alpha$ is an iteration coefficient, $\gamma$ is the learning rate, $\varepsilon_\theta$ means a noise prediction network, parameterized by $\theta$. $\mathcal{N}(0, \sigma^2I)$ is the added gaussian noise. 
In Eq. \eqref{diffusion-iterations}, the $\alpha$, $\gamma$, $\sigma$ are the functions of iteration step $k$ called noise schedule~\cite{nichol2021improved}. The underlying noise schedule controls the extent to which diffusion policy captures high and low-frequency characteristics of action signals. 
In the training process, we first randomly sample the original samples $x^0$ from the offline dataset. For each sample, we then randomly select a denoising iteration $k$, followed by randomly sampling noise $\varepsilon^k$ with an appropriate variance for iteration $k$. The loss function for the noise prediction network is as follows:
\begin{equation}
    \mathcal{L}_x = \mathrm{MSE}(\varepsilon^k, \varepsilon_\theta(x^0 + \varepsilon^k, k)).
    \label{noiseloss}
\end{equation}
To adapt the diffusion model for our task, we propose our GCAD method. Specifically, we modify the model output to be the robot actions with decreasing levels of noise $A_t^k, A_t^{k-1},..., A_t^0$ and change the input conditions for the denoising process by adding the observation values $O_t$ as computational conditions. 
In addition, we embed the return-to-go setting to the embedding layer to enable our method to learn optimal and robust operational actions with different action distributions. It provides a high-level task planning conditional constraint for the learned action trajectories to ensure the transformer action outputs move towards the suitable external structure for the current environment. 

According to the aforementioned, we use the action diffusion model with $\hat{R}_t$ embedding to approximate the conditional distribution $p(A_t | O_t, \hat{R}_t)$ instead of the joint distribution $p(A_t, O_t)$. This is done to enable the model to directly predict actions based on the current observations and the hint of return-to-go, without the need for predictions of future states. The improved formula is as follows:
\begin{equation}
    A_t^{k-1} = \alpha(A_t^k-\gamma\varepsilon_\theta(O_t, \hat{R}_t, A_t^k, k)+\mathcal{N}(0, \sigma^2I)),
    \label{ouriterations}
\end{equation}
On this basis, the loss function is as follows:
\begin{equation}
    \mathcal{L}_{O, A, \hat{R}} = \mathrm{MSE}(\varepsilon^k, \varepsilon_\theta(O_t, \hat{R}_t, A_t^0 + \varepsilon^k, k)).
    \label{ourloss}
\end{equation}
We only apply denoising prediction to actions, simplifying the inference process operationally. This simplification has a positive impact on predicting the entire robotic control sequence, allowing effective utilization of data with mixed visual inputs~\cite{diffusion}.

\section{Experiments}
In our experiments, we aim to answer the following questions: (1) Can our DexDiff method robustly grasp large and flat objects by extrinsic dexterity like humans? (2) Can our main contributors enhance the performance of task and motion planning? (3) Can our method generalize in unrestricted environments? (4) Can our method be deployed in real-world environments?

\begin{wrapfigure}{r}{0.3\textwidth}
    \centering
    \includegraphics[width=0.3\textwidth]{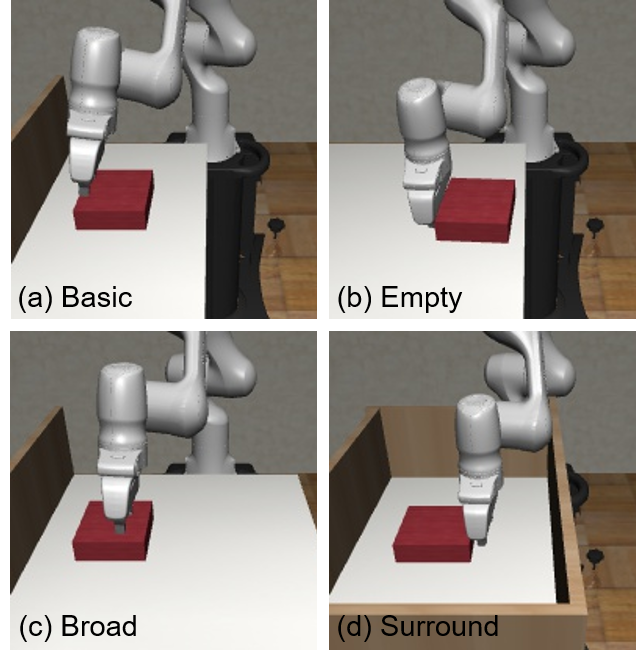}
    \caption{The simulation environments: (a) Basic, (b) Empty, (c) Broad, (d) Surround.}
    \label{simulation}
    \vspace{-0.4cm}
\end{wrapfigure}

\subsection{Experimental Setup and Metrics}
\textbf{Environments.} To evaluate DexDiff for the task of grasping flat objects with extrinsic dexterity, we build the basic simulation scenario and its several modification scenarios shown in Figure.~\ref{simulation} using the Robosuite and the Mujoco simulator~\cite{todorov2012mujoco, zhu2020robosuite}, namely: (a) \textit{Basic}, the object is lying flat on the table, with a supporting wall on the left side and the edge of the table on the right side, (b) \textit{Empty}, there is no supporting wall on the table, (c) \textit{Broad}, the table is very large and the table edge is far from the object, and (d) \textit{Surround}, the object is surrounded by supporting walls. Each experimental scenario contains a Franka Emika Panda robot positioned at one corner of a table.

\textbf{Datasets.} For task planning, we uniformly collected $\sim$ 1.6k images in the simulation scenarios and $\sim$ 1.4k images in the real scenarios, and provided a corpus of manipulation matched with corresponding scene images.
For motion planning, we construct trajectory datasets by combining reinforcement learning and human demonstration with 400 robot trajectories for each simulation scenario and 200 robot trajectories for each real-world scenario. Details of the dataset can be found in the Appendix~\ref{appendix_offline} and ~\ref{appendix_realworld}.

\begin{wrapfigure}{r}{0.55\textwidth}
    \centering
    \vspace{-0.3cm}
    \includegraphics[width=0.5\textwidth]{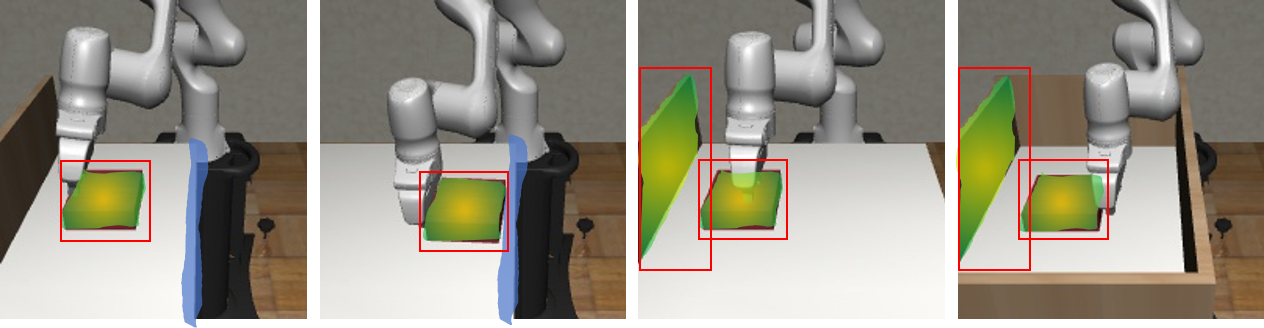}
    \vspace{0.4cm}
    \scriptsize
    \centering
    \begin{tabular}{c c c}
        \toprule
        \multicolumn{1}{c|}{Task Planners} & \multicolumn{1}{c|}{Motion Planners} & Avg.SR(\%) \\
        \midrule
        \multicolumn{1}{c|}{Heuristic} &  \multicolumn{1}{c|}{Behavior Cloning} & 0.00 \\
        \multicolumn{1}{c|}{Heuristic} &  \multicolumn{1}{c|}{Diffusion Policy-T} & 0.13 \\
        \multicolumn{1}{c|}{Heuristic} &  \multicolumn{1}{c|}{GCAD} & 0.35 \\
        \midrule
        \multicolumn{1}{c|}{Finetuned-VLM} & \multicolumn{1}{c|}{Behavior Cloning}  & 0.00 \\
        \multicolumn{1}{c|}{Finetuned-VLM} & \multicolumn{1}{c|}{Diffusion Policy-T}  & 0.23 \\
        \midrule
        \multicolumn{2}{c|}{\textbf{DexDiff(ours)}}  & \textbf{0.70} \\
        \midrule
    \end{tabular}
    \vspace{-0.4cm}
    \caption{Compared to traditional heuristic methods based on image segmentation and fixed rules.}
    \label{heuristic}
    \vspace{-0.4cm}
\end{wrapfigure}

\textbf{Baselines.} To validate the reasoning ability of the VLM module to make high-level decisions for task planning, we design a traditional heuristic method as a baseline for comparison (Figure.~\ref{heuristic}). The method uses Yolox~\cite{yolo} to detect the position of the object and the wall in the image, and the threshold method to detect the position of the table edge. Then the position of the wall as well as the table edge from the object is calculated separately. Finally, the external structure that is closest to the object is chosen as the high-level planning condition, which is used to guide the low-level operations. 

Moreover, we use the behavior clone (BC), Decision Transformer (DT)~\cite{decisiontransformer}, and Diffusion Policy as the low-level baseline~\cite{diffusion}. In detail, we compare our method to two variants of the Diffusion Policy: CNN-based Diffusion Policy (DP-C) and Transformer-based Diffusion Policy (DP-T). 

\textbf{Evaluation Metrics.} Our evaluation metrics for task completion are success rate and distance error. The success rate in the experiment is the proportion of successful samples to the total tests. We performed 35 tests per simulation scenario and 10 tests per real airfield scenario. The distance error $D(g, E)$ measures the weighted distance between the end effector $E$ and the desired grasping pose $g$ at the end of the task which is calculated from the translation distance $\Delta T(g, E)$ and rotation distance $\Delta R(g, E)$. It primarily evaluates the accuracy of the robot's actions. The specific form of the calculation is as follows:
\begin{equation}
    D(g, E) = \frac{1}{N}\sum ( \alpha_1 \Delta T(g, E) + \alpha_2 \Delta R(g, E) ).
    \label{distance}
\end{equation}
where the $\alpha_1$ and $\alpha_2$ represent the weight parameters of the distance error, and $N$ represents the total number of tests. We only apply this metric in the simulation.

\begin{table}
\scriptsize
\centering
\begin{tabular}{c c |c c| c c| c c| c c| c c}
    \toprule
    & & \multicolumn{2}{c|}{Basic} & \multicolumn{2}{c|}{Empty} & \multicolumn{2}{c|}{Broad} & \multicolumn{2}{c|}{Surround} & \multicolumn{2}{c}{Average}\\ 
    & & Dist. E & SR & Dist. E & SR & Dist. E & SR & Dist. E & SR & Dist. E & SR\\
    \midrule
    \multicolumn{2}{c|}{Finetuned-VLM \& BC} & -21.2 & - & -20.6 & - & -17.2 & - & -17.4 & - & -19.1 & -\\
    \multicolumn{2}{c|}{Finetuned-VLM \& DT~\cite{decisiontransformer}} & -9.71 & - & -9.72 & - & -15.6 & - & -17.2 & - & -13.1 & -\\
    \multicolumn{2}{c|}{Finetuned-VLM \& DP-C~\cite{diffusion}} & -3.99 & 0.03 & -3.67 & 0.14 & -4.88 & 0.14 & -4.67 & 0.09 & -4.30 & 0.10\\
    \multicolumn{2}{c|}{Finetuned-VLM \& DP-T~\cite{diffusion}} & -3.95 & 0.31 & -3.87 & 0.29 & -4.80 & 0.14 & -4.82 & 0.17 & -4.27 & 0.23\\
    \multicolumn{2}{c|}{\textbf{Finetuned-VLM \& GCAD (DexDiff)}} & \textbf{-1.93} & \textbf{0.85} & \textbf{-1.97} & \textbf{0.80} & \textbf{-3.24} & \textbf{0.63} & \textbf{-4.27} & \textbf{0.51} & \textbf{-2.85} & \textbf{0.70}\\
    \bottomrule
\end{tabular}
\vspace{0.2cm}
\caption{In simulation evaluations, our DexDiff demonstrates a higher success rate compared to baseline approaches across various scenarios and significantly minimizes distance error.}
\label{resulttable}
\vspace{-0.6cm}
\end{table}

\subsection{DexDiff Method Performance in Simulation}
To be fair, we use the VLM as the high-level planning module for all baseline methods. The experimental results in simulation environments are shown in Table.~\ref{resulttable}. Compared to baselines, the GCAD model achieves the highest average success rate of 70\% and the lowest distance error in the four scenarios.

Among the diffusion methods in the baselines, the transformer-based approach surpasses CNN, particularly in robot manipulation tasks requiring multi-modal inputs and high precision. This may be because CNNs tend to favor low-frequency signals, while extrinsic dexterity tasks require actions that change rapidly over time~\cite{diffusion}.
The BC and Decision Transformer methods fail to learn successful trajectories due to their limited effectiveness in long-horizon planning with multi-modal inputs and single-step action prediction.

GCAD, built on a Diffusion Policy network, excels in high-dimensional sequence modeling. By embedding the return-to-go within the network, the policy focuses on learning high-quality action data within high-quality trajectories. In summary, GCAD demonstrates superior learning capabilities for high precision and long-horizon planning in robot manipulations. Thus, our DexDiff method can robustly grasp large and flat objects by extrinsic dexterity by coordinating the VLM and GCAD models.


\subsection{Evaluation for Task Planners}
To validate the reasoning ability of the finetuned VLM module to make high-level decisions for task planning, we compare it with the traditional heuristics in several motion planning settings, and the results of the simulation experiments (Figure.~\ref{heuristic}) show that our approach has a higher average success rate of 70\%. This suggests that regardless of the motion planning method used, with our finetuned VLM model, it is possible to provide suitable and reliable plans for ungraspable tasks, which facilitates the selection of operational strategies.

\begin{wrapfigure}{r}{0.56\textwidth}
    \centering
    \vspace{-0.3cm}
    \includegraphics[width=0.56\textwidth]{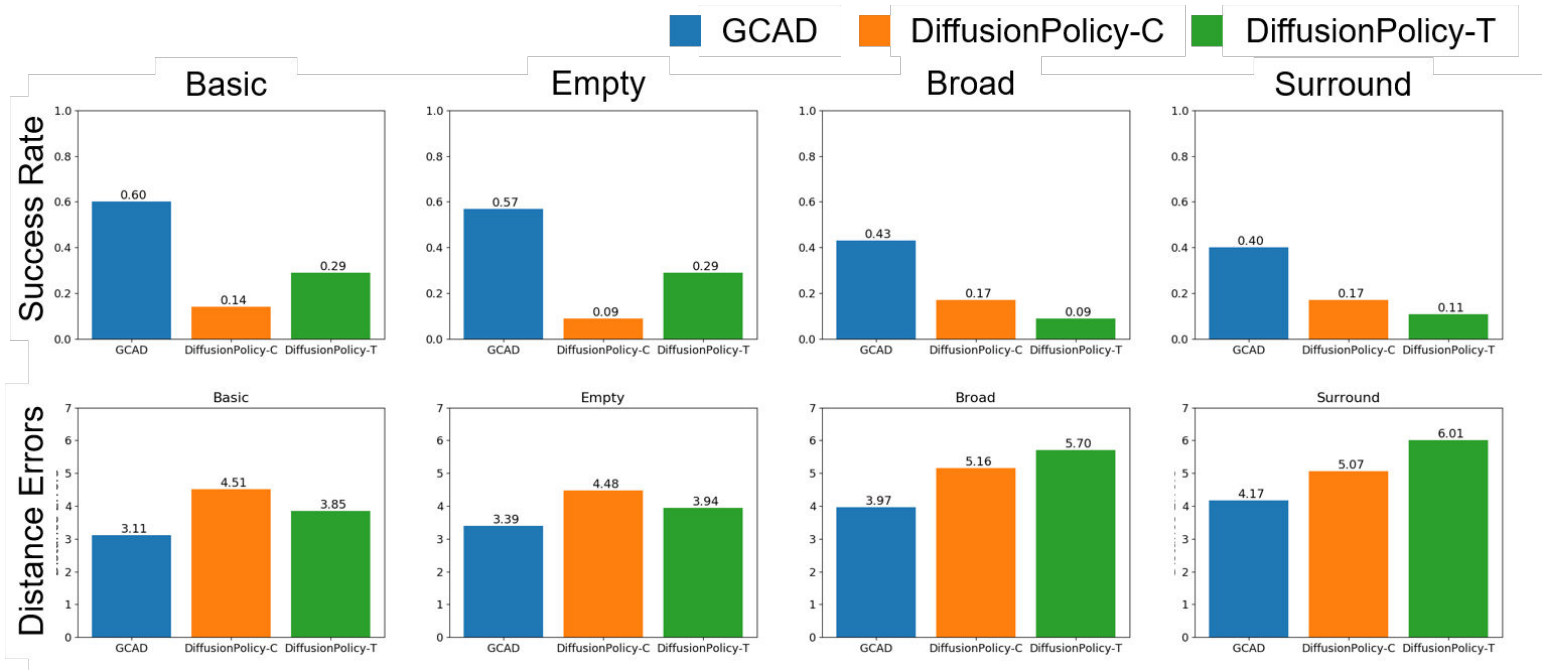}
    \caption{We evaluate the generalization ability of GCAD in different experiment settings.}
    \label{generalization}
    \vspace{-0.4cm}
\end{wrapfigure}

\subsection{Generalization in Simulation}
We validate the generalization ability of the models with settings outside the distribution of the datasets, by randomizing the position range, pose, size, and friction of the object, and the initial position of the gripper. Compared Table.~\ref{resulttable} and Figure.~\ref{generalization}, although our approach has some performance degradation, it still maintains the highest average success rate and the lowest distance error. We believe this is because the return-to-go added to the diffusion model of the transformer architecture guides the policy to learn and generalize, rather than simple imitation. 

\begin{figure}
    \centering
    \includegraphics[width= 0.95\textwidth]{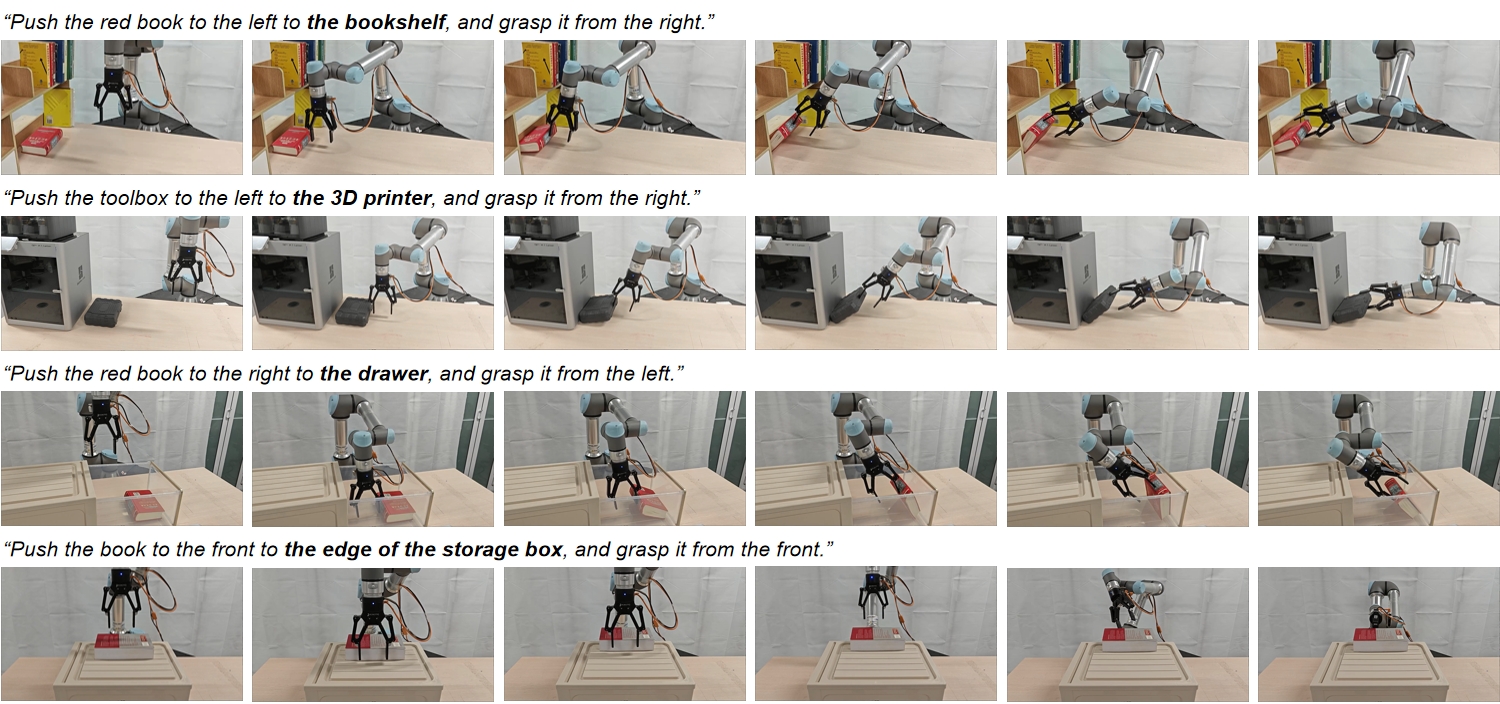}
    \caption{We use DexDiff to perform different manipulations with extrinsic dexterity in the real world.}
    \label{demo}
\end{figure}

\begin{table}
\centering
\scriptsize
\vspace{-0.2cm}
\begin{tabular}{c c c c c c}
    \toprule
     & \multicolumn{5}{c}{SR(\%)}\\
     & Bookshelf & 3D Printer (unseen) & Drawer & Storage box & Avg.\\ 
    \midrule
    Finetuned-VLM + BC &  0.00 (0/10) & 0.00 (0/10) & 0.00 (0/10) & 0.00 (0/10) & 0.00 (0/40)\\
    Heuristic + GCAD &  0.40 (4/10) & 0.30 (3/10) & 0.30 (3/10) & 0.40 (4/10) & 0.35 (14/40)\\
    \textbf{DexDiff (ours)} &  \textbf{0.70 (7/10)} & \textbf{0.50 (5/10)} & \textbf{0.60 (6/10)} & \textbf{0.80 (8/10)} & \textbf{0.65 (26/40)}\\
    \bottomrule
\end{tabular}
\vspace{0.2cm}
\caption{We evaluate DexDiff on the real robot with various daily-life scenarios, including unseen extrinsic dexterity.}
\vspace{-0.5cm}
\label{realresult1}
\end{table}

\subsection{Real-world Deployment}
We evaluate DexDiff in the real-world environment (Figure.~\ref{demo}). Our robot platform includes a UR5e robot with a Robotiq gripper and a front-view RGB camera. The experimental scenarios include retrieving a book from a bookshelf, grasping a toolbox with a 3D printer on a table, picking up a book from a drawer, and grasping a book on a storage box. Before the environment was set up, we did not inject fixed elements: there must be a standard table edge and a standard wall. Instead, we have guided the VLM model to understand the role of walls and tables in extrinsic dexterity through human priori knowledge. Our task planner can understand that the bookshelf, the drawer interior, and the side of the 3D printer can act as ``walls” in addition to the standard walls and the edge of the storage box can act as ``table edge” in addition to the standard table edge. 
We evaluate 10 cases for each test scenario and the results are shown in Table.~\ref{realresult1} (Note that the 3D printer environment is unseen). Compared to baselines, our method has better generalization ability in both task planning and motion planning parts and achieves a 65\% average success rate. 

\begin{wraptable}{r}{0.35\textwidth}
    \scriptsize
    \centering
    \vspace{-0.3cm}
    \begin{tabular}{c c c}
        \toprule
        Object-ID & Edge & Wall\\ 
        \midrule
        Box-0 & 6/10 & 3/10\\
        Box-1 (seen) & 8/10 & 6/10\\
        Box-2 & 8/10 & 4/10\\
        Toolbox (seen) & 7/10 & 5/10\\
        Folder & 9/10 & -\\
        Book & 6/10 & -\\
        Pan (seen) & 7/10 & -\\
        Medicine cabinet & 8/10 & -\\
        \midrule
        \textbf{Average SR} & 73.75\% & 45.0\%\\
        \bottomrule
    \end{tabular}
    \caption{We evaluate DexDiff on the real robot with various test objects, including some unseen samples.}
    \label{realresult}
    \vspace{-0.4cm}
\end{wraptable}

To further demonstrate the generalization ability of our method in ungraspable tasks, we evaluate the performance of the method for daily-life objects with different sizes, densities, and surface frictions in a standard environment (with a standard wall and edge of the table). The results are shown in Table.~\ref{realresult}. Our DexDiff method achieves a 73.75\% average success rate during the manipulation of push-to-table-edge and a 45\% average success rate during the operation of push-to-wall. This suggests that the learning of policies can be generalized to other shapes of objects in tasks with less demanding accuracy requirements. However, the push-to-wall manipulation requires high-precision actions. Any small error can lead to failure, especially the lack of friction on the side of the object.
The overall results show that our method can achieve a satisfactory success rate in the real world as well, and has practical usage value.


\section{Conclusion and Limitations}
\textbf{Conclusion.} In this paper, we introduced DexDiff, a robotic manipulation method designed for ungraspable problems with extrinsic dexterity. Through evaluations of extrinsic dexterity tasks in simulation and the real world, our goal-conditioned action diffusion model, guided by the high-level plans from the vision-language model, effectively executes the grasping task of large, flat objects robustly, consistently outperforming existing approaches. 
The experimental results also demonstrate that our method can understand and utilize non-standard external structures to perform extrinsic dexterity manipulation, and can be generalized to various unseen objects and real-world scenarios, showing practical value.


\textbf{Limitations.} Even though we have demonstrated the effectiveness of our robotic method in the ungraspable task, some limitations still exist.
In the environment perception and task planning module, whether the VLM can really understand the physical structure through visual input and make more intelligent high-level task planning based on lower-level motion planning remains an open question.
Due to transformer architecture, the GCAD model requires much demonstration. It is computationally expensive and the decision latency affects the smoothness of the real robot deployment. 


\clearpage


\bibliography{Dexdiff_arxiv}  

\begin{thebibliography}{70}
\providecommand{\natexlab}[1]{#1}
\providecommand{\url}[1]{\texttt{#1}}
\expandafter\ifx\csname urlstyle\endcsname\relax
  \providecommand{\doi}[1]{doi: #1}\else
  \providecommand{\doi}{doi: \begingroup \urlstyle{rm}\Url}\fi

\bibitem[Zhou and Held(2023)]{ungraspable}
W.~Zhou and D.~Held.
\newblock Learning to grasp the ungraspable with emergent extrinsic dexterity.
\newblock In \emph{Conference on Robot Learning}, pages 150--160. PMLR, 2023.

\bibitem[Dafle et~al.(2014)Dafle, Rodriguez, Paolini, Tang, Srinivasa, Erdmann, Mason, Lundberg, Staab, and Fuhlbrigge]{extrinsic}
N.~C. Dafle, A.~Rodriguez, R.~Paolini, B.~Tang, S.~S. Srinivasa, M.~Erdmann, M.~T. Mason, I.~Lundberg, H.~Staab, and T.~Fuhlbrigge.
\newblock Extrinsic dexterity: In-hand manipulation with external forces.
\newblock In \emph{2014 IEEE International Conference on Robotics and Automation (ICRA)}, pages 1578--1585. IEEE, 2014.

\bibitem[Sun et~al.(2020)Sun, Yuan, Hu, Yang, and Li]{sun2020learning}
Z.~Sun, K.~Yuan, W.~Hu, C.~Yang, and Z.~Li.
\newblock Learning pregrasp manipulation of objects from ungraspable poses.
\newblock In \emph{2020 IEEE International Conference on Robotics and Automation (ICRA)}, pages 9917--9923. IEEE, 2020.

\bibitem[Liang et~al.(2021)Liang, Lou, Yang, and Choi]{liang2021learning}
H.~Liang, X.~Lou, Y.~Yang, and C.~Choi.
\newblock Learning visual affordances with target-orientated deep q-network to grasp objects by harnessing environmental fixtures.
\newblock In \emph{2021 IEEE International Conference on Robotics and Automation (ICRA)}, pages 2562--2568. IEEE, 2021.

\bibitem[Zhang et~al.(2023)Zhang, Liang, Cong, Lyu, Zeng, Feng, and Zhang]{pag}
H.~Zhang, H.~Liang, L.~Cong, J.~Lyu, L.~Zeng, P.~Feng, and J.~Zhang.
\newblock Reinforcement learning based pushing and grasping objects from ungraspable poses.
\newblock \emph{IEEE International Conference on Robotics and Automation (ICRA)}, 2023.

\bibitem[Cheng et~al.(2023)Cheng, Patil, Temel, Kroemer, and Mason]{cheng2023enhancing}
X.~Cheng, S.~Patil, Z.~Temel, O.~Kroemer, and M.~T. Mason.
\newblock Enhancing dexterity in robotic manipulation via hierarchical contact exploration.
\newblock \emph{IEEE Robotics and Automation Letters}, 9\penalty0 (1):\penalty0 390--397, 2023.

\bibitem[Cheng et~al.(2021)Cheng, Huang, Hou, and Mason]{cheng2021contact}
X.~Cheng, E.~Huang, Y.~Hou, and M.~T. Mason.
\newblock Contact mode guided sampling-based planning for quasistatic dexterous manipulation in 2d.
\newblock In \emph{2021 IEEE International Conference on Robotics and Automation (ICRA)}, pages 6520--6526. IEEE, 2021.

\bibitem[Vaswani et~al.(2017)Vaswani, Shazeer, Parmar, Uszkoreit, Jones, Gomez, Kaiser, and Polosukhin]{transformer}
A.~Vaswani, N.~Shazeer, N.~Parmar, J.~Uszkoreit, L.~Jones, A.~N. Gomez, {\L}.~Kaiser, and I.~Polosukhin.
\newblock Attention is all you need.
\newblock \emph{Advances in neural information processing systems}, 30, 2017.

\bibitem[Zhou et~al.(2023)Zhou, Jiang, Yang, Paxton, and Held]{zhou2023hacman}
W.~Zhou, B.~Jiang, F.~Yang, C.~Paxton, and D.~Held.
\newblock Hacman: Learning hybrid actor-critic maps for 6d non-prehensile manipulation.
\newblock \emph{arXiv preprint arXiv:2305.03942}, 2023.

\bibitem[Kim et~al.(2023)Kim, Han, Kim, and Kim]{kim2023pre}
M.~Kim, J.~Han, J.~Kim, and B.~Kim.
\newblock Pre-and post-contact policy decomposition for non-prehensile manipulation with zero-shot sim-to-real transfer.
\newblock In \emph{2023 IEEE/RSJ International Conference on Intelligent Robots and Systems (IROS)}, pages 10644--10651. IEEE, 2023.

\bibitem[Cho et~al.(2024)Cho, Han, Cho, and Kim]{cho2024corn}
Y.~Cho, J.~Han, Y.~Cho, and B.~Kim.
\newblock Corn: Contact-based object representation for nonprehensile manipulation of general unseen objects.
\newblock \emph{arXiv preprint arXiv:2403.10760}, 2024.

\bibitem[Bogdanovic et~al.(2020)Bogdanovic, Khadiv, and Righetti]{bogdanovic2020learning}
M.~Bogdanovic, M.~Khadiv, and L.~Righetti.
\newblock Learning variable impedance control for contact sensitive tasks.
\newblock \emph{IEEE Robotics and Automation Letters}, 5\penalty0 (4):\penalty0 6129--6136, 2020.

\bibitem[Aeronautiques et~al.(1998)Aeronautiques, Howe, Knoblock, McDermott, Ram, Veloso, Weld, Sri, Barrett, Christianson, et~al.]{aeronautiques1998pddl}
C.~Aeronautiques, A.~Howe, C.~Knoblock, I.~D. McDermott, A.~Ram, M.~Veloso, D.~Weld, D.~W. Sri, A.~Barrett, D.~Christianson, et~al.
\newblock Pddl-the planning domain definition language.
\newblock \emph{Technical Report, Tech. Rep.}, 1998.

\bibitem[Kootbally et~al.(2015)Kootbally, Schlenoff, Lawler, Kramer, and Gupta]{kootbally2015towards}
Z.~Kootbally, C.~Schlenoff, C.~Lawler, T.~Kramer, and S.~K. Gupta.
\newblock Towards robust assembly with knowledge representation for the planning domain definition language (pddl).
\newblock \emph{Robotics and Computer-Integrated Manufacturing}, 33:\penalty0 42--55, 2015.

\bibitem[Thomason et~al.(2020)Thomason, Padmakumar, Sinapov, Walker, Jiang, Yedidsion, Hart, Stone, and Mooney]{thomason2020jointly}
J.~Thomason, A.~Padmakumar, J.~Sinapov, N.~Walker, Y.~Jiang, H.~Yedidsion, J.~Hart, P.~Stone, and R.~Mooney.
\newblock Jointly improving parsing and perception for natural language commands through human-robot dialog.
\newblock \emph{Journal of Artificial Intelligence Research}, 67:\penalty0 327--374, 2020.

\bibitem[Vallati et~al.(2015)Vallati, Chrpa, Grze{\'s}, McCluskey, Roberts, Sanner, et~al.]{vallati20152014}
M.~Vallati, L.~Chrpa, M.~Grze{\'s}, T.~L. McCluskey, M.~Roberts, S.~Sanner, et~al.
\newblock The 2014 international planning competition: Progress and trends.
\newblock \emph{Ai Magazine}, 36\penalty0 (3):\penalty0 90--98, 2015.

\bibitem[Huang et~al.(2022)Huang, Abbeel, Pathak, and Mordatch]{huang2022language}
W.~Huang, P.~Abbeel, D.~Pathak, and I.~Mordatch.
\newblock Language models as zero-shot planners: Extracting actionable knowledge for embodied agents.
\newblock In \emph{International Conference on Machine Learning}, pages 9118--9147. PMLR, 2022.

\bibitem[Brohan et~al.(2023)Brohan, Chebotar, Finn, Hausman, Herzog, Ho, Ibarz, Irpan, Jang, Julian, et~al.]{saycan}
A.~Brohan, Y.~Chebotar, C.~Finn, K.~Hausman, A.~Herzog, D.~Ho, J.~Ibarz, A.~Irpan, E.~Jang, R.~Julian, et~al.
\newblock Do as i can, not as i say: Grounding language in robotic affordances.
\newblock In \emph{Conference on robot learning}, pages 287--318. PMLR, 2023.

\bibitem[Wang et~al.(2023)Wang, Cai, Chen, Liu, Ma, and Liang]{wang2023describe}
Z.~Wang, S.~Cai, G.~Chen, A.~Liu, X.~Ma, and Y.~Liang.
\newblock Describe, explain, plan and select: Interactive planning with large language models enables open-world multi-task agents.
\newblock \emph{arXiv preprint arXiv:2302.01560}, 2023.

\bibitem[Vemprala et~al.(2024)Vemprala, Bonatti, Bucker, and Kapoor]{chatgptforrobot}
S.~H. Vemprala, R.~Bonatti, A.~Bucker, and A.~Kapoor.
\newblock Chatgpt for robotics: Design principles and model abilities.
\newblock \emph{IEEE Access}, 2024.

\bibitem[Liang et~al.(2023)Liang, Huang, Xia, Xu, Hausman, Ichter, Florence, and Zeng]{liang2023code}
J.~Liang, W.~Huang, F.~Xia, P.~Xu, K.~Hausman, B.~Ichter, P.~Florence, and A.~Zeng.
\newblock Code as policies: Language model programs for embodied control.
\newblock In \emph{2023 IEEE International Conference on Robotics and Automation (ICRA)}, pages 9493--9500. IEEE, 2023.

\bibitem[Brohan et~al.(2022)Brohan, Brown, Carbajal, Chebotar, Dabis, Finn, Gopalakrishnan, Hausman, Herzog, Hsu, et~al.]{rt1}
A.~Brohan, N.~Brown, J.~Carbajal, Y.~Chebotar, J.~Dabis, C.~Finn, K.~Gopalakrishnan, K.~Hausman, A.~Herzog, J.~Hsu, et~al.
\newblock Rt-1: Robotics transformer for real-world control at scale.
\newblock \emph{arXiv preprint arXiv:2212.06817}, 2022.

\bibitem[Zitkovich et~al.(2023)Zitkovich, Yu, Xu, Xu, Xiao, Xia, Wu, et~al.]{rt2}
B.~Zitkovich, T.~Yu, S.~Xu, P.~Xu, T.~Xiao, F.~Xia, J.~Wu, et~al.
\newblock {RT}-2: Vision-language-action models transfer web knowledge to robotic control.
\newblock In \emph{7th Annual Conference on Robot Learning}, 2023.
\newblock URL \url{https://openreview.net/forum?id=XMQgwiJ7KSX}.

\bibitem[Li et~al.(2023)Li, Liu, Zhang, Yu, Xu, Wu, Cheang, Jing, Zhang, Liu, et~al.]{li2023vision}
X.~Li, M.~Liu, H.~Zhang, C.~Yu, J.~Xu, H.~Wu, C.~Cheang, Y.~Jing, W.~Zhang, H.~Liu, et~al.
\newblock Vision-language foundation models as effective robot imitators.
\newblock \emph{arXiv preprint arXiv:2311.01378}, 2023.

\bibitem[Padalkar et~al.(2023)Padalkar, Pooley, Jain, Bewley, Herzog, Irpan, Khazatsky, Rai, Singh, Brohan, et~al.]{padalkar2023open}
A.~Padalkar, A.~Pooley, A.~Jain, A.~Bewley, A.~Herzog, A.~Irpan, A.~Khazatsky, A.~Rai, A.~Singh, A.~Brohan, et~al.
\newblock Open x-embodiment: Robotic learning datasets and rt-x models.
\newblock \emph{arXiv preprint arXiv:2310.08864}, 2023.

\bibitem[Tellex et~al.(2020)Tellex, Gopalan, Kress-Gazit, and Matuszek]{tellex2020robots}
S.~Tellex, N.~Gopalan, H.~Kress-Gazit, and C.~Matuszek.
\newblock Robots that use language.
\newblock \emph{Annual Review of Control, Robotics, and Autonomous Systems}, 3:\penalty0 25--55, 2020.

\bibitem[Gao et~al.(2023)Gao, Sarkar, Xia, Xiao, Wu, Ichter, Majumdar, and Sadigh]{gao2023physically}
J.~Gao, B.~Sarkar, F.~Xia, T.~Xiao, J.~Wu, B.~Ichter, A.~Majumdar, and D.~Sadigh.
\newblock Physically grounded vision-language models for robotic manipulation.
\newblock \emph{arXiv preprint arXiv:2309.02561}, 2023.

\bibitem[Driess et~al.(2023)Driess, Xia, Sajjadi, Lynch, Chowdhery, Ichter, Wahid, Tompson, Vuong, Yu, et~al.]{palm}
D.~Driess, F.~Xia, M.~S. Sajjadi, C.~Lynch, A.~Chowdhery, B.~Ichter, A.~Wahid, J.~Tompson, Q.~Vuong, T.~Yu, et~al.
\newblock Palm-e: An embodied multimodal language model.
\newblock \emph{arXiv preprint arXiv:2303.03378}, 2023.

\bibitem[Du et~al.(2023)Du, Yang, Florence, Xia, Wahid, Ichter, Sermanet, Yu, Abbeel, Tenenbaum, et~al.]{VLP}
Y.~Du, M.~Yang, P.~Florence, F.~Xia, A.~Wahid, B.~Ichter, P.~Sermanet, T.~Yu, P.~Abbeel, J.~B. Tenenbaum, et~al.
\newblock Video language planning.
\newblock \emph{arXiv preprint arXiv:2310.10625}, 2023.

\bibitem[Khandate et~al.(2023)Khandate, Shang, Chang, Saidi, Adams, and Ciocarlie]{Khandate-RSS-23}
G.~Khandate, S.~Shang, E.~T. Chang, T.~L. Saidi, J.~Adams, and M.~Ciocarlie.
\newblock {Sampling-based Exploration for Reinforcement Learning of Dexterous Manipulation}.
\newblock In \emph{Proceedings of Robotics: Science and Systems}, Daegu, Republic of Korea, July 2023.
\newblock \doi{10.15607/RSS.2023.XIX.020}.

\bibitem[Zhang et~al.(2023)Zhang, Ke, Deshpande, Gupta, and Srinivasa]{Zhang-RSS-23}
Y.~Zhang, L.~Ke, A.~Deshpande, A.~Gupta, and S.~Srinivasa.
\newblock {Cherry-Picking with Reinforcement Learning}.
\newblock In \emph{Proceedings of Robotics: Science and Systems}, Daegu, Republic of Korea, July 2023.
\newblock \doi{10.15607/RSS.2023.XIX.021}.

\bibitem[Li et~al.(2023)Li, Peng, Abbeel, Levine, Berseth, and Sreenath]{Li-RSS-23}
Z.~Li, X.~B. Peng, P.~Abbeel, S.~Levine, G.~Berseth, and K.~Sreenath.
\newblock {Robust and Versatile Bipedal Jumping Control through Reinforcement Learning}.
\newblock In \emph{Proceedings of Robotics: Science and Systems}, Daegu, Republic of Korea, July 2023.
\newblock \doi{10.15607/RSS.2023.XIX.052}.

\bibitem[Huang et~al.(2022)Huang, Hu, and Jayaraman]{huang2022training}
K.~Huang, E.~S. Hu, and D.~Jayaraman.
\newblock Training robots to evaluate robots: Example-based interactive reward functions for policy learning.
\newblock In \emph{6th Annual Conference on Robot Learning}, 2022.
\newblock URL \url{https://openreview.net/forum?id=sK2aWU7X9b8}.

\bibitem[Cruciani et~al.(2018)Cruciani, Smith, Kragic, and Hang]{cruciani2018dexterous}
S.~Cruciani, C.~Smith, D.~Kragic, and K.~Hang.
\newblock Dexterous manipulation graphs.
\newblock In \emph{2018 IEEE/RSJ International Conference on Intelligent Robots and Systems (IROS)}, pages 2040--2047. IEEE, 2018.

\bibitem[Cruciani et~al.(2019)Cruciani, Hang, Smith, and Kragic]{cruciani2019dualarm}
S.~Cruciani, K.~Hang, C.~Smith, and D.~Kragic.
\newblock Dual-arm in-hand manipulation and regrasping using dexterous manipulation graphs, 2019.

\bibitem[Zeng et~al.(2018)Zeng, Song, Welker, Lee, Rodriguez, and Funkhouser]{zeng2018learning}
A.~Zeng, S.~Song, S.~Welker, J.~Lee, A.~Rodriguez, and T.~Funkhouser.
\newblock Learning synergies between pushing and grasping with self-supervised deep reinforcement learning.
\newblock In \emph{2018 IEEE/RSJ International Conference on Intelligent Robots and Systems (IROS)}, pages 4238--4245. IEEE, 2018.

\bibitem[Xu et~al.(2021)Xu, Yu, Lai, Wang, and Xiong]{xu2021efficient}
K.~Xu, H.~Yu, Q.~Lai, Y.~Wang, and R.~Xiong.
\newblock Efficient learning of goal-oriented push-grasping synergy in clutter.
\newblock \emph{IEEE Robotics and Automation Letters}, 6\penalty0 (4):\penalty0 6337--6344, 2021.

\bibitem[Haldar et~al.(2023)Haldar, Pari, Rai, and Pinto]{Haldar-RSS-23}
S.~Haldar, J.~Pari, A.~Rai, and L.~Pinto.
\newblock {Teach a Robot to FISH: Versatile Imitation from One Minute of Demonstrations}.
\newblock In \emph{Proceedings of Robotics: Science and Systems}, Daegu, Republic of Korea, July 2023.
\newblock \doi{10.15607/RSS.2023.XIX.009}.

\bibitem[Li et~al.(2023)Li, Keipour, Jamieson, Hudson, Swan, and Bekris]{Li2-RSS-23}
S.~Li, A.~Keipour, K.~Jamieson, N.~Hudson, C.~Swan, and K.~Bekris.
\newblock {Demonstrating Large-Scale Package Manipulation via Learned Metrics of Pick Success}.
\newblock In \emph{Proceedings of Robotics: Science and Systems}, Daegu, Republic of Korea, July 2023.
\newblock \doi{10.15607/RSS.2023.XIX.023}.

\bibitem[Liu et~al.(2023)Liu, Nasiriany, Zhang, Bao, and Zhu]{Liu-RSS-23}
H.~Liu, S.~Nasiriany, L.~Zhang, Z.~Bao, and Y.~Zhu.
\newblock {Robot Learning on the Job: Human-in-the-Loop Autonomy and Learning During Deployment}.
\newblock In \emph{Proceedings of Robotics: Science and Systems}, Daegu, Republic of Korea, July 2023.
\newblock \doi{10.15607/RSS.2023.XIX.005}.

\bibitem[Zeng et~al.(2023)Zeng, Ichter, Xia, Xiao, and Sindhwani]{Zeng-RSS-23}
A.~Zeng, B.~Ichter, F.~Xia, T.~Xiao, and V.~Sindhwani.
\newblock {Demonstrating Large Language Models on Robots}.
\newblock In \emph{Proceedings of Robotics: Science and Systems}, Daegu, Republic of Korea, July 2023.
\newblock \doi{10.15607/RSS.2023.XIX.024}.

\bibitem[Schuppe et~al.(2023)Schuppe, Torre, Leite, and Tumova]{Schuppe-RSS-23}
G.~Schuppe, I.~Torre, I.~Leite, and J.~Tumova.
\newblock {Follow my Advice: Assume-Guarantee Approach to Task Planning with Human in the Loop}.
\newblock In \emph{Proceedings of Robotics: Science and Systems}, Daegu, Republic of Korea, July 2023.
\newblock \doi{10.15607/RSS.2023.XIX.001}.

\bibitem[Kostrikov et~al.(2023)Kostrikov, Smith, and Levine]{Kostrikov-RSS-23}
I.~Kostrikov, L.~M. Smith, and S.~Levine.
\newblock {Demonstrating A Walk in the Park: Learning to Walk in 20 Minutes With Model-Free Reinforcement Learning}.
\newblock In \emph{Proceedings of Robotics: Science and Systems}, Daegu, Republic of Korea, July 2023.
\newblock \doi{10.15607/RSS.2023.XIX.056}.

\bibitem[Zhang et~al.(2018)Zhang, McCarthy, Jow, Lee, Chen, Goldberg, and Abbeel]{zhang2018deep}
T.~Zhang, Z.~McCarthy, O.~Jow, D.~Lee, X.~Chen, K.~Goldberg, and P.~Abbeel.
\newblock Deep imitation learning for complex manipulation tasks from virtual reality teleoperation.
\newblock In \emph{2018 IEEE International Conference on Robotics and Automation (ICRA)}, pages 5628--5635. IEEE, 2018.

\bibitem[Mandlekar et~al.(2020)Mandlekar, Xu, Mart{\'\i}n-Mart{\'\i}n, Savarese, and Fei-Fei]{mandlekar2020learning}
A.~Mandlekar, D.~Xu, R.~Mart{\'\i}n-Mart{\'\i}n, S.~Savarese, and L.~Fei-Fei.
\newblock Learning to generalize across long-horizon tasks from human demonstrations.
\newblock \emph{arXiv preprint arXiv:2003.06085}, 2020.

\bibitem[Zeng et~al.(2021)Zeng, Florence, Tompson, Welker, Chien, Attarian, Armstrong, Krasin, Duong, Sindhwani, et~al.]{zeng2021transporter}
A.~Zeng, P.~Florence, J.~Tompson, S.~Welker, J.~Chien, M.~Attarian, T.~Armstrong, I.~Krasin, D.~Duong, V.~Sindhwani, et~al.
\newblock Transporter networks: Rearranging the visual world for robotic manipulation.
\newblock In \emph{Conference on Robot Learning}, pages 726--747. PMLR, 2021.

\bibitem[Mandlekar et~al.(2020)Mandlekar, Ramos, Boots, Savarese, Fei-Fei, Garg, and Fox]{mandlekar2020iris}
A.~Mandlekar, F.~Ramos, B.~Boots, S.~Savarese, L.~Fei-Fei, A.~Garg, and D.~Fox.
\newblock Iris: Implicit reinforcement without interaction at scale for learning control from offline robot manipulation data.
\newblock In \emph{2020 IEEE International Conference on Robotics and Automation (ICRA)}, pages 4414--4420. IEEE, 2020.

\bibitem[Chen et~al.(2021)Chen, Lu, Rajeswaran, Lee, Grover, Laskin, Abbeel, Srinivas, and Mordatch]{decisiontransformer}
L.~Chen, K.~Lu, A.~Rajeswaran, K.~Lee, A.~Grover, M.~Laskin, P.~Abbeel, A.~Srinivas, and I.~Mordatch.
\newblock Decision transformer: Reinforcement learning via sequence modeling.
\newblock \emph{Advances in neural information processing systems}, 34:\penalty0 15084--15097, 2021.

\bibitem[Xu et~al.(2022)Xu, Shen, Zhang, Lu, Zhao, Tenenbaum, and Gan]{xu2022prompting}
M.~Xu, Y.~Shen, S.~Zhang, Y.~Lu, D.~Zhao, J.~Tenenbaum, and C.~Gan.
\newblock Prompting decision transformer for few-shot policy generalization.
\newblock In \emph{international conference on machine learning}, pages 24631--24645. PMLR, 2022.

\bibitem[Ajay et~al.(2022)Ajay, Du, Gupta, Tenenbaum, Jaakkola, and Agrawal]{ajay2022conditional}
A.~Ajay, Y.~Du, A.~Gupta, J.~Tenenbaum, T.~Jaakkola, and P.~Agrawal.
\newblock Is conditional generative modeling all you need for decision-making?
\newblock \emph{arXiv preprint arXiv:2211.15657}, 2022.

\bibitem[Ho et~al.(2020)Ho, Jain, and Abbeel]{ho2020denoising}
J.~Ho, A.~Jain, and P.~Abbeel.
\newblock Denoising diffusion probabilistic models.
\newblock \emph{Advances in neural information processing systems}, 33:\penalty0 6840--6851, 2020.

\bibitem[Yoneda et~al.(2023)Yoneda, Sun, Yang, Stadie, and Walter]{Yoneda-RSS-23}
T.~Yoneda, L.~Sun, G.~Yang, B.~C. Stadie, and M.~R. Walter.
\newblock {To the Noise and Back: Diffusion for Shared Autonomy}.
\newblock In \emph{Proceedings of Robotics: Science and Systems}, Daegu, Republic of Korea, July 2023.
\newblock \doi{10.15607/RSS.2023.XIX.014}.

\bibitem[Janner et~al.(2022)Janner, Du, Tenenbaum, and Levine]{janner2022planning}
M.~Janner, Y.~Du, J.~Tenenbaum, and S.~Levine.
\newblock Planning with diffusion for flexible behavior synthesis.
\newblock In \emph{International Conference on Machine Learning}, pages 9902--9915. PMLR, 2022.

\bibitem[Liu et~al.(2023)Liu, Du, Hermans, Chernova, and Paxton]{Liu2-RSS-23}
W.~Liu, Y.~Du, T.~Hermans, S.~Chernova, and C.~Paxton.
\newblock {StructDiffusion: Language-Guided Creation of Physically-Valid Structures using Unseen Objects}.
\newblock In \emph{Proceedings of Robotics: Science and Systems}, Daegu, Republic of Korea, July 2023.
\newblock \doi{10.15607/RSS.2023.XIX.031}.

\bibitem[Reuss et~al.(2023)Reuss, Li, Jia, and Lioutikov]{Reuss-RSS-23}
M.~Reuss, M.~Li, X.~Jia, and R.~Lioutikov.
\newblock {Goal-Conditioned Imitation Learning using Score-based Diffusion Policies}.
\newblock In \emph{Proceedings of Robotics: Science and Systems}, Daegu, Republic of Korea, July 2023.
\newblock \doi{10.15607/RSS.2023.XIX.028}.

\bibitem[Huang et~al.(2023)Huang, Wang, Li, Jia, Liu, Zhu, Liang, and Zhu]{huang2023diffusion}
S.~Huang, Z.~Wang, P.~Li, B.~Jia, T.~Liu, Y.~Zhu, W.~Liang, and S.-C. Zhu.
\newblock Diffusion-based generation, optimization, and planning in 3d scenes.
\newblock In \emph{Proceedings of the IEEE/CVF Conference on Computer Vision and Pattern Recognition}, pages 16750--16761, 2023.

\bibitem[Wang et~al.(2022)Wang, Hunt, and Zhou]{wang2022diffusion}
Z.~Wang, J.~J. Hunt, and M.~Zhou.
\newblock Diffusion policies as an expressive policy class for offline reinforcement learning.
\newblock \emph{arXiv preprint arXiv:2208.06193}, 2022.

\bibitem[Chi et~al.(2023)Chi, Feng, Du, Xu, Cousineau, Burchfiel, and Song]{diffusion}
C.~Chi, S.~Feng, Y.~Du, Z.~Xu, E.~Cousineau, B.~Burchfiel, and S.~Song.
\newblock Diffusion policy: Visuomotor policy learning via action diffusion.
\newblock \emph{Robotics: Science and Systems (RSS)}, 2023.

\bibitem[Song and Ermon(2019)]{NEURIPS2019_3001ef25}
Y.~Song and S.~Ermon.
\newblock Generative modeling by estimating gradients of the data distribution.
\newblock In H.~Wallach, H.~Larochelle, A.~Beygelzimer, F.~d\textquotesingle Alch\'{e}-Buc, E.~Fox, and R.~Garnett, editors, \emph{Advances in Neural Information Processing Systems}, volume~32. Curran Associates, Inc., 2019.
\newblock URL \url{https://proceedings.neurips.cc/paper_files/paper/2019/file/3001ef257407d5a371a96dcd947c7d93-Paper.pdf}.

\bibitem[Du et~al.(2021)Du, Qian, Liu, Ding, Qiu, Yang, and Tang]{du2021glm}
Z.~Du, Y.~Qian, X.~Liu, M.~Ding, J.~Qiu, Z.~Yang, and J.~Tang.
\newblock Glm: General language model pretraining with autoregressive blank infilling.
\newblock \emph{arXiv preprint arXiv:2103.10360}, 2021.

\bibitem[Xu et~al.(2024)Xu, Zhang, Li, Liu, Lan, and Kong]{xu2024sinvig}
J.~Xu, H.~Zhang, X.~Li, H.~Liu, X.~Lan, and T.~Kong.
\newblock Sinvig: A self-evolving interactive visual agent for human-robot interaction.
\newblock \emph{arXiv preprint arXiv:2402.11792}, 2024.

\bibitem[Nichol and Dhariwal(2021)]{nichol2021improved}
A.~Q. Nichol and P.~Dhariwal.
\newblock Improved denoising diffusion probabilistic models.
\newblock In \emph{International Conference on Machine Learning}, pages 8162--8171. PMLR, 2021.

\bibitem[Todorov et~al.(2012)Todorov, Erez, and Tassa]{todorov2012mujoco}
E.~Todorov, T.~Erez, and Y.~Tassa.
\newblock Mujoco: A physics engine for model-based control.
\newblock In \emph{2012 IEEE/RSJ international conference on intelligent robots and systems}, pages 5026--5033. IEEE, 2012.

\bibitem[Zhu et~al.(2020)Zhu, Wong, Mandlekar, Mart{\'\i}n-Mart{\'\i}n, Joshi, Nasiriany, and Zhu]{zhu2020robosuite}
Y.~Zhu, J.~Wong, A.~Mandlekar, R.~Mart{\'\i}n-Mart{\'\i}n, A.~Joshi, S.~Nasiriany, and Y.~Zhu.
\newblock robosuite: A modular simulation framework and benchmark for robot learning.
\newblock \emph{arXiv preprint arXiv:2009.12293}, 2020.

\bibitem[Redmon et~al.(2016)Redmon, Divvala, Girshick, and Farhadi]{yolo}
J.~Redmon, S.~Divvala, R.~Girshick, and A.~Farhadi.
\newblock You only look once: Unified, real-time object detection.
\newblock In \emph{Proceedings of the IEEE conference on computer vision and pattern recognition}, pages 779--788, 2016.

\bibitem[He et~al.(2020)He, Fan, Wu, Xie, and Girshick]{he2020momentum}
K.~He, H.~Fan, Y.~Wu, S.~Xie, and R.~Girshick.
\newblock Momentum contrast for unsupervised visual representation learning.
\newblock In \emph{Proceedings of the IEEE/CVF conference on computer vision and pattern recognition}, pages 9729--9738, 2020.

\bibitem[Wu and He(2018)]{wu2018group}
Y.~Wu and K.~He.
\newblock Group normalization.
\newblock In \emph{Proceedings of the European conference on computer vision (ECCV)}, pages 3--19, 2018.

\bibitem[Mandlekar et~al.(2021)Mandlekar, Xu, Wong, Nasiriany, Wang, Kulkarni, Fei-Fei, Savarese, Zhu, and Mart{\'\i}n-Mart{\'\i}n]{mandlekar2021matters}
A.~Mandlekar, D.~Xu, J.~Wong, S.~Nasiriany, C.~Wang, R.~Kulkarni, L.~Fei-Fei, S.~Savarese, Y.~Zhu, and R.~Mart{\'\i}n-Mart{\'\i}n.
\newblock What matters in learning from offline human demonstrations for robot manipulation.
\newblock \emph{arXiv preprint arXiv:2108.03298}, 2021.

\bibitem[Mayne and Michalska(1988)]{mayne1988receding}
D.~Q. Mayne and H.~Michalska.
\newblock Receding horizon control of nonlinear systems.
\newblock In \emph{Proceedings of the 27th IEEE Conference on Decision and Control}, pages 464--465. IEEE, 1988.

\bibitem[Florence et~al.(2022)Florence, Lynch, Zeng, Ramirez, Wahid, Downs, Wong, Lee, Mordatch, and Tompson]{florence2022implicit}
P.~Florence, C.~Lynch, A.~Zeng, O.~A. Ramirez, A.~Wahid, L.~Downs, A.~Wong, J.~Lee, I.~Mordatch, and J.~Tompson.
\newblock Implicit behavioral cloning.
\newblock In \emph{Conference on Robot Learning}, pages 158--168. PMLR, 2022.

\end{thebibliography}
\clearpage
\appendix
\section{Appendix}
\label{appendix}
\subsection{Offline Datasets for Motion Planning in Simulation}\label{appendix_offline}
We collected different types of data sets in four simulation scenarios, namely Basic, Empty, Borad, and Surround, each simulation scenario collected 400 trajectories. In the Basic and Empty datasets, to achieve the highest operation success rate, the expected strategy is first to push the object to the edge of the table and then grasp it from the side. In the other two datasets, the optimal strategy is to push the object against the wall, lift it up, and then grasp it.
The horizon length of the Basic and Empty dataset is 70, and the horizon length of the Broad and Surround dataset is 40.

The method of collecting datasets is mainly based on the Soft Actor-Critic reinforcement learning algorithm, which guides the learning robot's policy network by setting different reward functions for different tasks. The observation includes the target grasp pose in the object frame, the gripper pose in the object frame, the object pose in the world frame and the gripper pose in the world frame. The robot's action is the 6-D motion of the gripper in 3-D space. The rewards for the Broad and Surround dataset mainly include the distance between the object and the table, the distance between the actual position of the gripper and the gripper target position, and the degree to which the target position of the gripper is blocked by the table and the wall. 

The collection method of the Basic and Empty datasets is mainly realized by learning two SAC networks. The first policy network guides the robot to push the object to the edge of the table. The reward function mainly includes the distance between the object block and the edge of the table. The second policy network guides the gripper to approach the target position of the gripper from a specified position, and the reward mainly includes the distance between the actual position of the gripper and the target position.
In order to prevent the robot from pushing objects too much along the edge of the table, we set boundary rewards so that the robot can push objects as much as possible in the direction perpendicular to the edge of the table. 
The entire action process is divided into three stages. The first stage involves the robot executing the first policy network in the first 10 timesteps of view to push the object to the edge of the table. The middle 20 timesteps of view in the second stage guide the robot from the end position of the first stage to the designated position. The last 40 timesteps of view in the third stage execute the second policy network to guide the robot's gripper to the target position of the gripper. Some important training parameters are shown in Table.~\ref{details}.

\begin{table}[h]
\centering
\begin{tabular}{c c c c c c c c c c}
    \toprule
     & horizon & obs image & gripper pose & action dim & $T_o$ & $T_a$ & lr & $\alpha_1$ & $\alpha_2$\\ 
    \midrule
    Basic &  70 & 256$\times$256$\times$3 & 7 & 6 & 2 & 2 & 1e-4 & 25 & 4 \\
    Empty &  70 & 256$\times$256$\times$3 & 7 & 6 & 2 & 2 & 1e-4 & 25 & 4 \\
    Broad &  40 & 256$\times$256$\times$3 & 7 & 6 & 2 & 2 & 1e-4 & 25 & 4 \\
    Surround &  40 & 256$\times$256$\times$3 & 7 & 6 & 2 & 2 & 1e-4 & 25 & 4 \\
    \bottomrule
\end{tabular}
\vspace{0.2cm}
\caption{Setting details in the simulation experiment.}
\label{details}
\end{table}

\subsection{Visualization of Simulation and Real-world Results}
\begin{figure}[t]
    \centering
    \includegraphics[width= \textwidth]{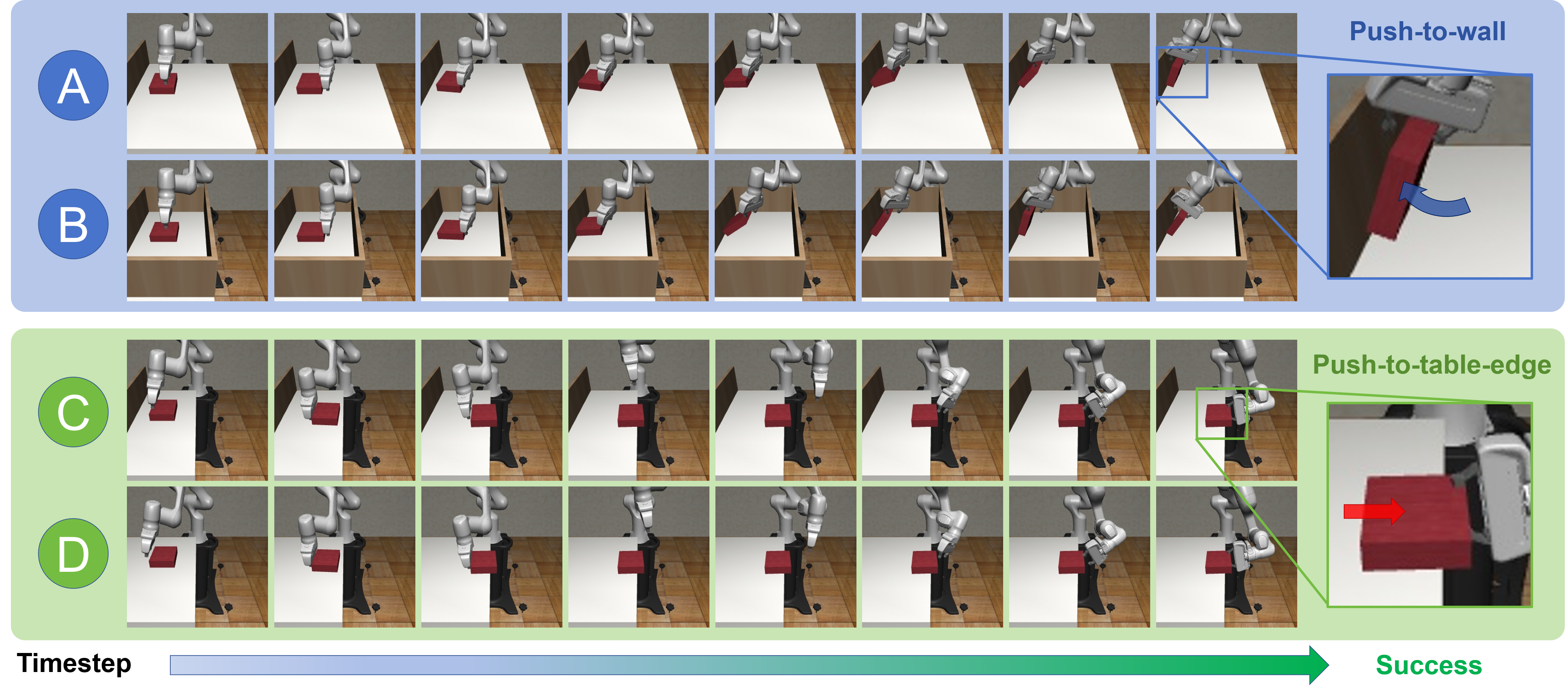}
    \caption{\textbf{A Visualization of Rollouts on the Simulation}. Depending on extrinsic dexterity structures, we get different high-level plans, and the proposed method enables precise robot manipulation on long-horizon tasks. (A) Broad, (B) Surround, (C) Basic, (D) Empty.}
    \label{resultdemo}
\end{figure}

We supplement visualizations of the simulation experiments of the DexDiff method (Figure.~\ref{resultdemo}). For the Broad and Surround environments, high-level planning often recommends push-to-wall as the pre-grasping action, while for the Basic and Empty environments, it tends to push the object to the edge of the table. Leveraging human-like task planning, the GCAD method can more effectively learn suitable actions, thereby accomplishing the task of grasping large and flat objects using extrinsic dexterity.

\begin{figure}[t]
    \centering
    \includegraphics[width= \textwidth]{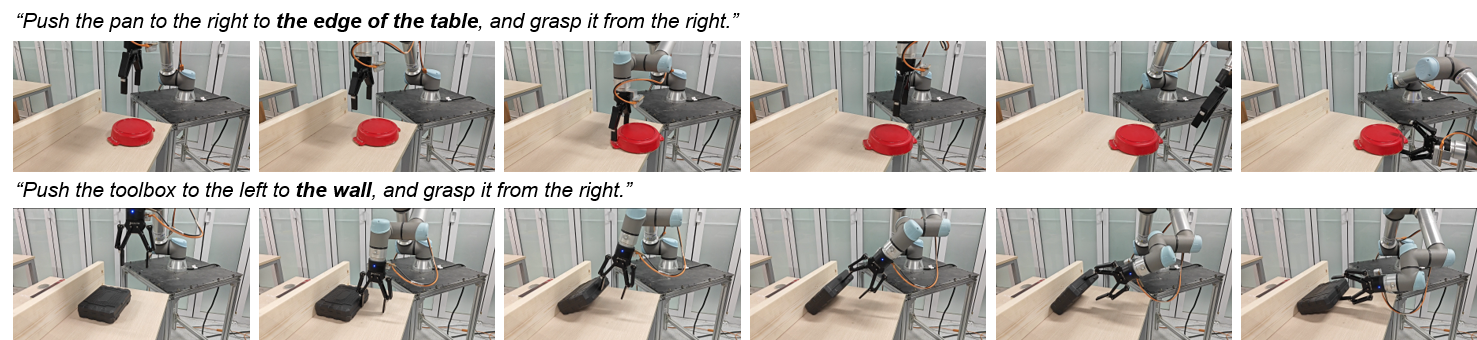}
    \caption{We use DexDiff to perform different manipulations with extrinsic dexterity in the real world.}
    \label{standard_demo}
\end{figure}

We complement this with a visual demonstration of the DexDiff method in real-world experiments(Figure.~\ref{standard_demo}). For toolboxes close to walls, advanced planning usually suggests pushing the toolbox against the wall as a pre-grabbing action. Since the curved sides of the pan make it difficult for the robot to lift one side of it using friction and extrinsic dexterity on the wall, high-level planning favors pushing the pan to the edge of the table and then grasping from the other side. In the motion planning module, we use the GCAD method to be able to learn appropriate motions more efficiently, thus enabling the task of grasping large flat objects using extrinsic dexterity.

\subsection{Realworld Details}\label{appendix_realworld}
\begin{figure}
    \centering
    \includegraphics[width= 3.2 in]{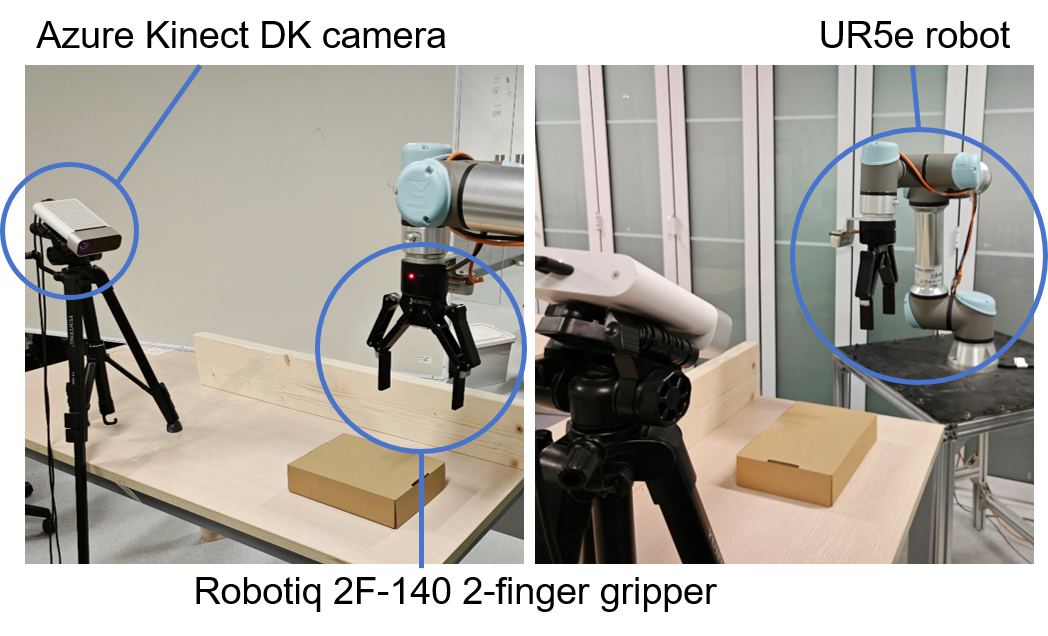}
    \caption{\textbf{Robot Platform Setup}. We evaluated the real-world performance of DexDiff using a UR5e robotic platform. }
    \label{robotplatform}
\end{figure}

The robot setup is shown in Figure.~\ref{robotplatform}. Our experimental scenario is a large flat object placed on a table, with the upper surface of the table at the same height as the robot base, and a fixed solid wall placed on the side of the object away from the edge of the table. We use the UR5e robot to train the robot to complete pre-grasping and final grasping actions. The camera uses the Azure Kinect DK camera to collect pictures in real time. The gripper that interacts with the object uses the Robotiq 2F-140 parallel gripper.

In collecting real-world data for the task planning module, we collected image data for four different positions of wall and table edges in the standard experimental scenarios, and in three near-daily-life scenarios. We collect a total of $\sim$1.4k images in real scenarios and provide a corpus of operations matching the images of the corresponding scenarios.

\begin{figure}
    \centering
    \includegraphics[width= 4 in]{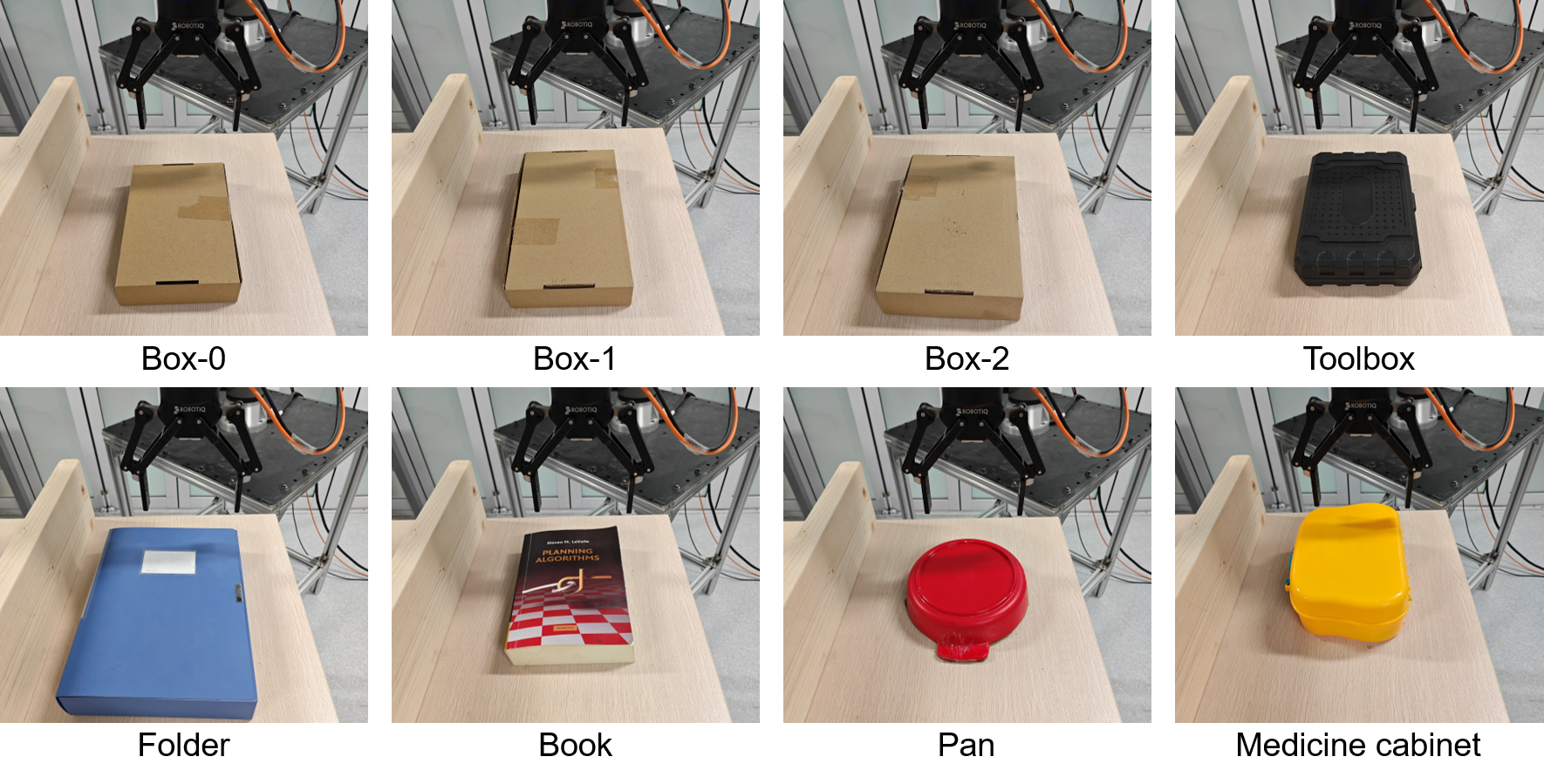}
    \caption{\textbf{Object generalization}. We evaluated the generalization performance of DexDiff using objects of different sizes and shapes. }
    \label{robotobject}
\end{figure}

For the motion planning module, we used a human-observed teaching-based approach to collect expert data and collect 200 trajectories for each scenario. Specifically, we set up demonstration expert trajectories for each setting of each environment, collect RGB images and robot gripper poses for the whole trajectory as states using a frequency of 2Hz, take the displacement of the gripper between every two steps as the robot's action, and set a sparse reward of 1 when the object is successfully grasped, and -1 otherwise. after 1500 epochs of training, the robot is used to test the success rate of different types of objects(Figure.~\ref{robotobject}) as well as scene settings.



\subsection{Visual Encoder and Noise Schedule}
We use a ResNet-18 as the visual encoder with some improvements (a spatial softmax pooling and the GroupNorm)~\cite{he2020momentum, wu2018group, mandlekar2021matters, diffusion}. This visual encoder maps the front-view RGB image sequence into an input embedding $O_t$ to train an action diffusion policy(encode the image independently for each time step). 

The noise schedule includes the Gaussian noise $\varepsilon^k$ and parameters $\alpha$, $\gamma$, $\theta$ in Eq.~\eqref{ouriterations}. The choice of noise schedule significantly influences the training of the denoising diffusion model policy, primarily because these noise parameters govern the model's ability to learn high-frequency and low-frequency features of the action signals, directly impacting our action diffusion. In robot control experiments, the Square Cosine Schedule has been demonstrated to be both simple and effective~\cite{nichol2021improved}. Therefore, our approach directly utilizes the Square Cosine Schedule as the noise schedule.

\subsection{Receding Horizon Control}
The complex process of predicting robot motion sequences involves long-horizon planning, especially for our ungraspable task. To maintain the accuracy of action learning in long-horizon planning, we refer to the receding horizontal control method~\cite{mayne1988receding}. The Goal-conditioned Action Diffusion model defines finite sequence steps for the diffusion model observations $O_t$, returns $\hat{R}_t$, and output actions $A_t$. Concretely, we choose an observation horizon $T_o$ and input the latest $T_o$ steps of observations $O_t$ and return-to-go values $\hat{R}_t$ at time step $t$. Subsequently, we predict the actions for the action horizon $T_a$ steps and execute the predicted actions on robots without re-planning. 

The design of action sequence prediction in this manner has several advantages. (1) It ensures the coherence and temporal consistency of the predicted actions. Long-horizon action planning requires continuous smooth control. If each action in the sequence is predicted as an independent distribution, the actions sampled and learned in different trajectories may produce a jitter when executing an action sequence continuously, especially for high-precision tasks with high cumulative errors~\cite{decisiontransformer}.
(2) It avoids the problem of mimicking high-frequency action distributions, which is common in single-step prediction. Actions that occur more frequently in the trajectory are often generated by waiting and delaying while data collection, which is manifested as a period of idle manipulation in the action sequence. Single-step policies make it easier to learn such actions and often tend to fit idle actions while ignoring the detailed actions that really matter. In contrast, the transformer model of sequence prediction that we employ can somewhat address the lack of attention to detailed actions.


\begin{figure}
    \centering
    \includegraphics[width=4 in]{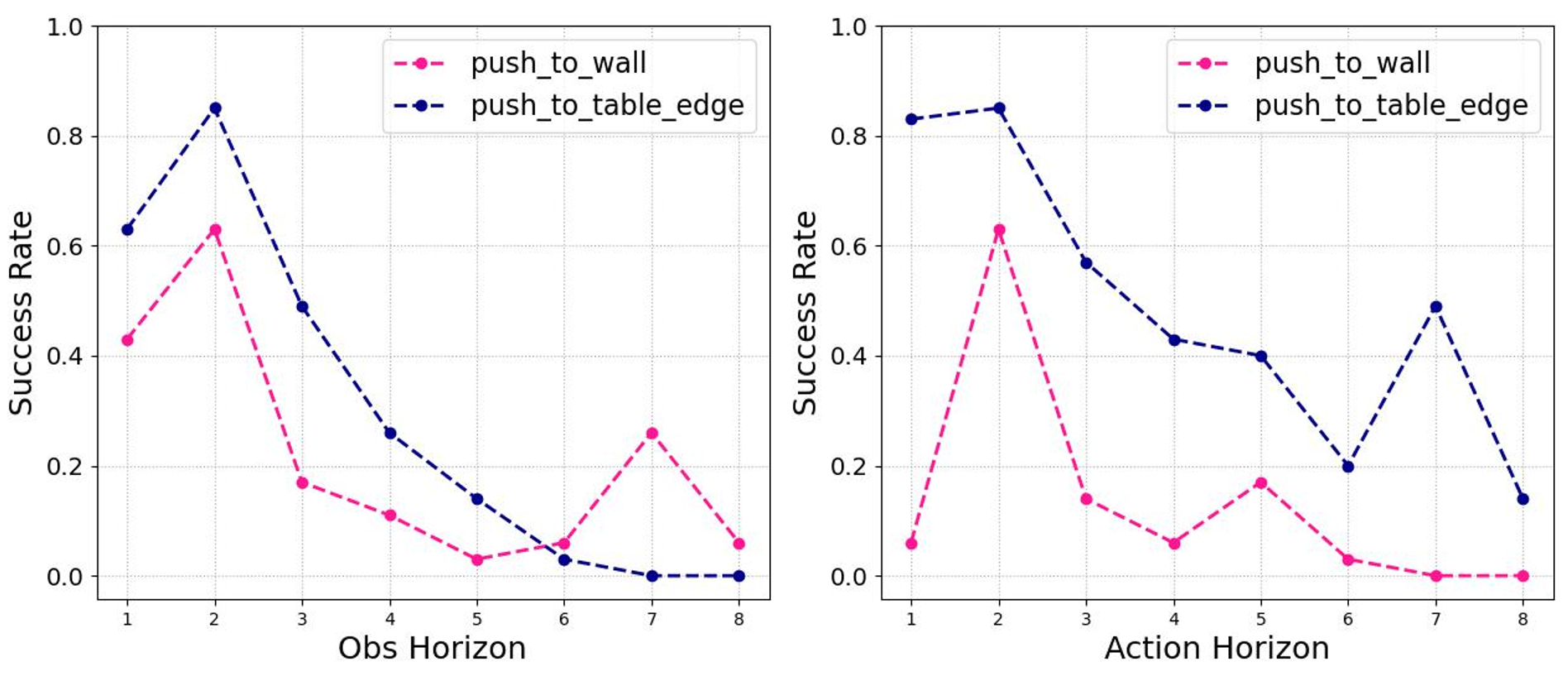}
    \caption{\textbf{Parameter Study curves.} [Left] observation horizon, [Right] action prediction horizon. Choosing the different lengths of observation horizon or action horizon significantly affects action prediction.}
    \label{xiaorong}
\end{figure}

\subsection{Parameter Study for Sequence Prediction}
Many policy learning methods avoid predicting action sequences because effectively sampling from high-dimensional output spaces poses significant challenges~\cite{florence2022implicit, mandlekar2021matters}. The diffusion model expands the dimensionality of the output without sacrificing the expressive power~\cite{diffusion}. Using this property, diffusion models can predict future actions in the form of high-dimensional action sequences. In addition, to learn the environmental information more continuously, we perform the input of observations in the form of sequences, and the potential relationship between the input images can also significantly affect the effectiveness.

The selection of the action prediction horizon $T_a$ and the observation horizon $T_o$ is related to the complexity, coherence, and sequence distribution of actions in specific work scenarios. Our parameter study confirms this tradeoff (Figure.~\ref{xiaorong}) and finds the two-step action prediction horizon and observation horizon to be optimal for our extrinsic dexterity tasks. A long action horizon helps to predict a sequence of consecutive valid actions, but too long a horizon leads to a higher cumulative error in prediction and is not conducive to task success. A long observation horizon provides more input information for building the mapping, but too long a horizon interferes with the accuracy of the action prediction at the current moment in time.

\subsection{Failure Case Analysis in Simulation}
\begin{figure}[h]
    \centering
    \includegraphics[width=\textwidth]{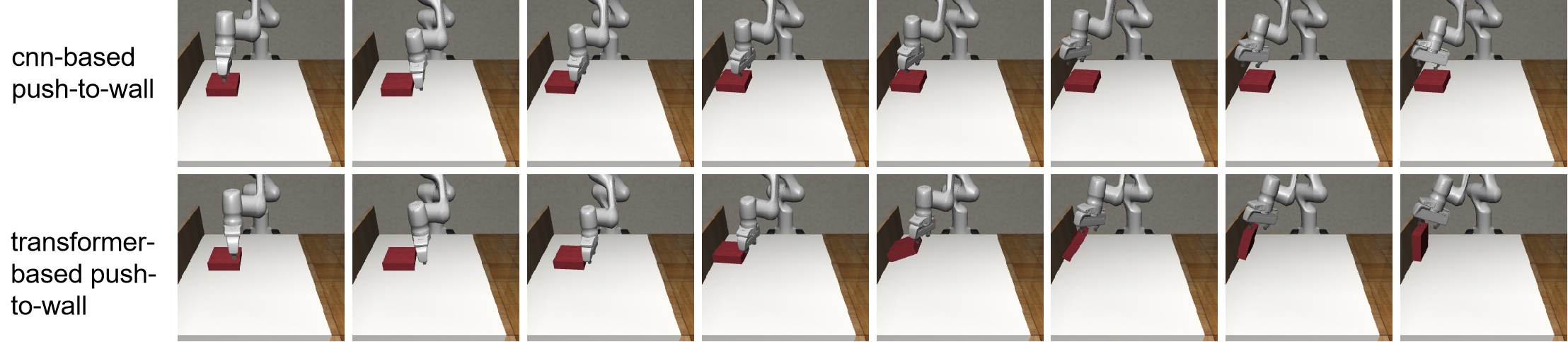}
    \caption{Baseline failure cases in simulation(Broad).  }
    \label{fail1}
    \centering
    \includegraphics[width=\textwidth]{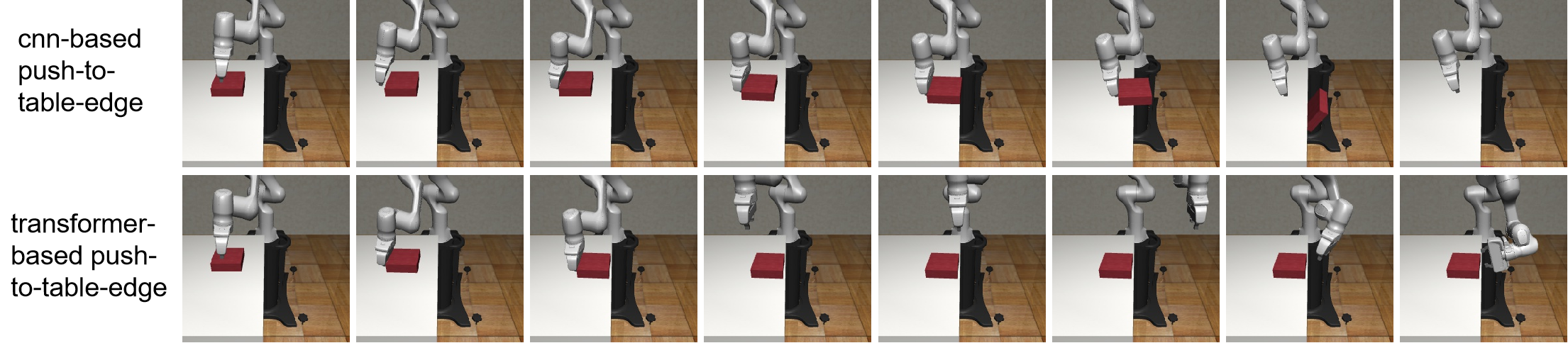}
    \caption{Baseline failure cases in simulation(Empty).}
    \label{fail2}
\end{figure}

We trained our dataset on traditional transformer-based diffusion and CNN-based diffusion and used the dataset for testing. The testing results of pushing the object against the wall are shown in Figure.~\ref{fail1}. 

We can see that when using cnn-base diffusion to train the robot to push the object to the wall and lift it up for grasping, the robot always lifts the gripper prematurely before pushing the object to the wall, causing the object not to be grasped. Lift, making grasping action impossible. We also see that when the transformer-based diffusion training robot was used to push the object to the wall and then lift it to grasp it, the robot always lifted the object to a too-high position, causing the entire object to lean against the wall. Due to the target grasping pose of the object being blocked by the wall, the robot is unable to grasp the object from above.
The testing results of pushing the object to the table edge and then grasping it are shown in Figure.~\ref{fail2}. 

We can see that when using CNN-based diffusion to train a robot to push an object to the edge of the table and then grasp the object from the side, it is easy to push the object too much, causing the object to be pushed to the ground by the robot. We also see that when a transformer-based diffusion training robot is used to push an object to the table, and the trained robot policy network is used to guide the robot to push the object, the pushing amplitude is easily too small, resulting in no part of the object being vacated, and therefore the robot's gripper can't grasping the object at the appropriate gripping position.

\subsection{Failure Case Analysis in the Real World}

\begin{figure}[h]
    \centering
    \includegraphics[width= \textwidth]{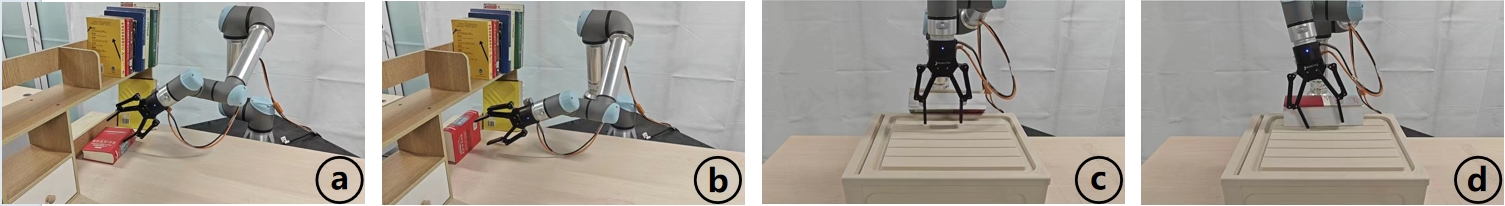}
    \caption{Failure cases in the real world.}
    \label{fail3}
\end{figure}

We also show some failure case demonstrations of our DexDiff approach in real scenario experiments shown in Figure.~\ref{fail3}. Since fine-grained positional control is required for quasi-static operation in our task, once the robot undergoes fine manipulation during operation, it may cause the following two types of failures: (a) as the robot rotates the book prematurely, the friction force based on which the robot supports the overhanging side of the book is drastically reduced, resulting in the book falling; and (b) when the robot supports the book lifting up by using friction force, it lifts up the book side too high, resulting in the book leaning on the bookshelf on its upright side and thus unable to further grasp the target object from above. 

Other types of failure occur when grasping from the side. Different types of failures happen when the robot grasps the book on the top surface of the storage box, due to the unavoidable loss of information that is difficult for the neural network to avoid when encoding the visual input image, the following failure scenarios may occur: (c) The step of the robot pushing the book operation is too long, resulting in the book falling off the storage box. (d) The height perception error of the robot operation causes the end-effector to collide with the top of the storage box.

\subsection{Different Shapes of the Side of the Table}

We made homemade irregular and curved edges fixed to table edges for experiments, to investigate the generalization performance and sensitivity of our DexDiff method for irregularly shaped edges. We used the model parameters of the standard push-table-edge to deploy directly in the environments. The experiments show that our method can be somewhat robust in the VLM task planning part, and performs some success cases in motion planning  (20\% (2/10) in irregular edge and 30\% (3/10) in curved edge).

The main reason for the failure may be the limitation of the trained model for the perception of distance in the visual input, which leads to the possibility of falling to the ground when performing pushing operations, or the gripper may hit the irregular edge when performing grasping operations.

\begin{figure}[h]
    \centering
    \includegraphics[width=4 in]{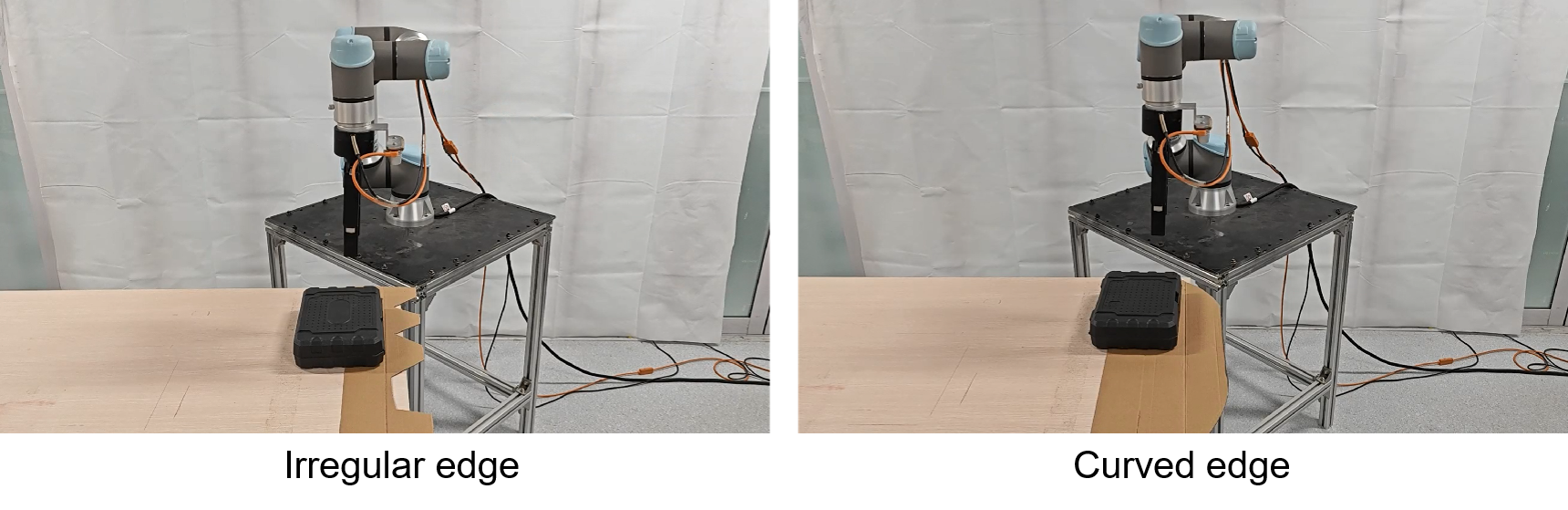}
    \caption{Different shapes of the side of the table.}
    \label{edge}
\end{figure}

\end{document}